\documentclass[
]{ceurart}

\usepackage{listings}

\usepackage{orcidlink}

\begin{document}

\conference{}

\title{The Future of Continual Learning in the Era of Foundation Models: Three Key Directions}

\author[1]{Jack Bell}[%
orcid=0009-0002-7857-4752,
email=jack.bell@di.unipi.it,
]
\cormark[1]
\address[1]{Department of Computer Science, Università di Pisa, 56126 Pisa, Italy}
\address[2]{Institute of Information Science and Technologies, National Research Council, 56124 Pisa, Italy}

\author[1]{Luigi Quarantiello}[%
orcid=0009-0005-5428-156X,
email=luigi.quarantiello@phd.unipi.it,
]

\author[1]{Eric Nuertey Coleman}[%
orcid=0009-0003-1319-9274,
email=eric.coleman@phd.unipi.it,
]

\author[1,2]{Lanpei Li}[%
orcid=0009-0005-4370-1020,
email=lanpei.li@isti.cnr.it,
]

\author[1]{Malio Li}[%
orcid=0009-0002-9572-7546,
email=malio.li@phd.unipi.it,
]

\author[1]{Mauro Madeddu}[%
orcid=0009-0002-7844-3963,
email=mauro.madeddu@phd.unipi.it,]

\author[1]{Elia Piccoli}[%
orcid=0009-0007-8923-3634,
email=elia.piccoli@phd.unipi.it,
]

\author[1]{Vincenzo Lomonaco}[%
orcid=0000-0001-8308-6599,
email=vincenzo.lomonaco@unipi.it,
]

\cortext[1]{Corresponding author.}

\begin{abstract}
Continual learning—the ability to acquire, retain, and refine knowledge over time—has always been fundamental to intelligence, both human and artificial. Historically, different AI paradigms have acknowledged this need, albeit with varying priorities: early expert and production systems focused on incremental knowledge consolidation, while reinforcement learning emphasised dynamic adaptation. With the rise of deep learning, deep continual learning has primarily focused on learning robust and reusable representations over time to solve sequences of increasingly complex tasks. However, the emergence of Large Language Models (LLMs) and foundation models has raised the question: Do we still need continual learning when centralised, monolithic models can tackle diverse tasks with access to internet-scale knowledge? We argue that continual learning remains essential for three key reasons: (i) continual pre-training is still necessary to ensure foundation models remain up to date, mitigating knowledge staleness and distribution shifts while integrating new information; (ii) continual fine-tuning enables models to specialise and personalise, adapting to domain-specific tasks, user preferences, and real-world constraints without full retraining, avoiding the need for computationally expensive long context-windows; (iii) continual compositionality offers a scalable and modular approach to intelligence, enabling the orchestration of foundation models and agents to be dynamically composed, recombined, and adapted. While continual pre-training and fine-tuning are explored as niche research directions, we argue it is continual compositionality that will mark the rebirth of continual learning. The future of AI will not be defined by a single static model but by an ecosystem of continually evolving and interacting models, making continual learning more relevant than ever.
\end{abstract}


\maketitle

\section{Introduction}

In recent years, artificial intelligence (AI) systems have begun to surpass human performance in many domains such as natural language processing (NLP) and computer vision. However, these models are typically static in nature and do not naturally update their understanding as new data emerges over time. In contrast, humans tend to approach problems as sequential learning tasks, building on past information without forgetting previously learned knowledge or requiring rehearsal to retain it \cite{french_catastrophic_1999}. Both human and AI systems require the ability to learn and adapt continuously, whilst avoiding so-called catastrophic forgetting, in which new learning erases previous knowledge. Addressing this challenge is one of the core aims of Continual Learning (CL).

Continual Learning research therefore revolves around two primary goals: \textbf{adaptation} and \textbf{memory consolidation}. Adaptation emphasises rapid responsiveness, enabling an agent to quickly adjust its behaviour or representations to maximise a utility function given the current task, situation or environment \cite{ditzler_2015}. Memory consolidation, on the other hand, involves building durable, generalisable knowledge and skills from past experiences deemed relevant to future tasks. This consolidation goes beyond mere retention; it focuses on developing abstract and hierarchical representations of knowledge, reusable across increasingly complex tasks over time. 

Historically, these two goals have been addressed with varying degrees of importance in different research methodologies, contexts and communities. Early expert systems, for instance, focused on consolidating incremental domain knowledge, yet lacked flexibility in rapidly adapting to new information without significant manual effort. Reinforcement learning (RL) methods, such as CHILD, introduced by Ring \cite{ring_continual_1994}, aimed instead at quick progressive learning, tackling easier tasks before addressing more complex ones, and adapting in a general approach towards continual reinforcement learning. Subsequent work on low-dimensional streaming data looked to address the issue of learning concept drifts \cite{ditzler_2015}, similarly focusing on rapid adaptation. With the steep rise and success of deep learning around 2012, the focus shifted prominently towards memory consolidation and generalisation. Deep continual learning leveraged neural networks to learn hierarchical, abstract representations directly from data, enabling the effective reuse of these representations across new tasks \cite{kumar_2012, giannini_2024}. Here, consolidation is not simply memory retention but involves the generalisation of latent knowledge and skills that facilitate adaptation to novel scenarios. We further explore the history of continual learning and its evolution to where it is today in section \ref{background}.

More recently, the emergence of LLMs has changed the focus of AI research towards transformer-based models with less focus on 'traditional' Machine Learning (ML). These models, pre-trained on vast datasets, have demonstrated remarkable capabilities to learn rich generalisations of the world \cite{li_2024}, performing well across a range of tasks. Combining these impressive capabilities, a demonstrated reduced propensity for catastrophic forgetting \cite{cossu_continual_2024} and access to the internet-scale knowledge, it is tempting to ask: \textit{Is continual learning still necessary within the era of foundation models?}

Applying continual learning, with its two main aims of adaptability and memory consolidation, to Foundation Models (FMs) is a way to overcome some of their inherent shortcomings: since a FM's parameters are fixed at deployment, every model can be seen as a snapshot of the world at the point of training. Practical use of FMs demands \textit{post-training} adaptation, through fine-tuning or personalisation for a downstream task. Therefore, their static nature poses a significant challenge --- they lack the intrinsic adaptability required to stay current in rapidly changing environments. 

So far, FM-based agents that actively interact with their environment have been proposed as a promising solution, leveraging continual adaptation to progressively improve capabilities \cite{zheng_2025}. However, the ability to adapt to new tasks is not enough; it is instead necessary to consolidate new knowledge over time to improve the overall understanding of the world. Challenges such as distributional shifts, long task sequences, task heterogeneity and inaccessible upstream data \cite{shi_continual_2024} necessitate a renewed focus on continual learning. Shi \textit{et al.} \cite{shi_continual_2024} further observe that these hurdles have pushed recent work toward \textit{task-incremental} and \textit{domain-incremental} benchmarks, where the task identity is supplied or irrelevant. Consequently, while such settings simplify experimentation for academic settings, real-world deployments still confront \textit{class-incremental} conditions in which entirely new tasks must be detected and learned on the fly \cite{van_de_ven_2022}. We further detail the need for CL in foundation models in section \ref{need-for-CL}.

Given this context, CL for foundational models is developing along three directions: \textbf{Continual Pre-Training (CPT)}, \textbf{Continual Fine-Tuning (CFT)} and \textbf{Continual Compositionality \& Orchestration (CCO)}. Human perception provides an instructive analogy: during infancy, critical periods enable rapid specialisation and consolidation of broad sensory capabilities such as vision and language discrimination \cite{hensch_2005}. Similarly, in ML, extensive pre-training ideally establishes a general-purpose foundational model, but CPT still remains practically necessary. Continual fine-tuning, in turn, allows efficient specialisation and personalisation to specific downstream tasks or contexts. However, both CPT and CFT typically require relatively lower-frequency adaptation cycles. Furthermore, their reliance on large-scale datasets, substantial computational resources and incremental improvement constraints due to scaling laws limits their potential to drive substantial advances alone \cite{shi_continual_2024}, particularly as these approaches predominantly extend established capabilities rather than enabling fundamentally new behaviours. However, recent work on {test-time scaling laws} demonstrates that, once model size passes a certain threshold, allocating additional inference-time compute delivers larger accuracy gains than further parameter growth \cite{chen_expanding_2025}. Parallel advances in multi-step chain-of-thought prompting and in multi-agent frameworks, where several specialised LLMs negotiate, critique, or divide labour, likewise point to performance improvements that arise from \textit{coordination} rather than monolithic scale \cite{schmidgall_agentrxiv_2025, cai2023human}. Together, these trends expose the practical limits of an end-to-end foundation model and highlight the need for a modular, dynamically reconfigurable approach. 

We therefore contend that \textbf{Continual Compositionality and Orchestration} represents the most promising and necessary direction for future continual learning research. Unlike CPT and CFT, CCO inherently supports high-frequency adaptation, allowing dynamic orchestration, recombination and collaborative interaction among multiple FMs or agents. Recent advances in FMs have primarily emerged not from additional computational resources, but rather through enhanced reasoning abilities \cite{kumar_2025} and longer context windows \cite{team2024gemini}, both of which are forms of orchestration rather than scale. Consequently, the future of continual learning likely resides in decentralised ecosystems, where multiple adaptive agents continuously interact, evolve and collaboratively address increasingly complex problems. This paradigm exhibits parallels to society as a whole, where collections of different individuals can come together to solve difficult tasks. Taken from a large enough intelligent population, a random sample of people will outperform a sample of the best performing agents --- with the intuition being that diversity is more important than individual ability \cite{hong2004groups}. In a similar vein, a diverse subset of agents may well be more adept at problem solving than a sample of the best performing agents at a given task. 

In this paper, we first review Continual Pre-Training (\ref{CPT}), which equips large-scale foundational models with adaptive, resource-efficient mechanisms to incorporate new knowledge without catastrophic forgetting. We then examine Continual Fine-Tuning (\ref{continual-fine-tuning}), enabling precise specialisation to downstream tasks while retaining broad, generalisable representations. Finally, we argue that Continual Compositionality \& Orchestration (\ref{continual-compositionality}) - with its high-frequency, modular coordination of specialised agents - offers the most promising path forward. By moving from monolithic snapshots to dynamic, decentralised ecosystems of models, CCO can drive the next wave of resilient, scalable, and sustainable AI systems.

\section{Background and Related Work}
\label{background}

The rise of Deep Learning (DL) in 2012 marked a pivotal moment for the entire ML research community. Initial efforts in this domain concentrated on utilising deep neural networks for representation learning, allowing models to capture abstract and hierarchical features from data. Nevertheless, when trained sequentially on multiple tasks, these models remained susceptible to catastrophic forgetting. This highlights the need for CL, which has had a large impact on the broader field of ML.

A widely used definition, often taken to specify Deep Continual Learning, is offered by Lesort \textit{et al.} \cite{lesort_continual_2019}. They describe continual learning as a learning paradigm where a model learns from a continuous stream of data, adapting to new information while preserving previously acquired knowledge. This definition emphasises the importance of both stability and plasticity in the learning process, where there is an important trade-off between retaining past knowledge and being plastic enough to adapt to new data or domains. The objective of CL is to have a machine learning model that can be adapted quickly to shifts in data distribution or ``tasks'', enabling it to retain already acquired knowledge and concepts and reuse these representations to facilitate better learning across new tasks.

This is a key difference between CL and traditional machine learning approaches, which typically require retraining on a static dataset to incorporate new information.

Early studies focused on the problem of catastrophic forgetting in neural networks \cite{mccloskey_catastrophic_1989, ratcliff1990connectionist}, where authors discovered the degradation of model performance on previous tasks while learning a new one. To overcome the issue, different basic approaches were proposed \cite{srivastava_compete_2013, goodfellow_empirical_2015, li_learning_2017,kirkpatrick_overcoming_2017}. These early works laid the foundations for more sophisticated approaches that emerged in the following years.

The main approaches within CL can be categorised into three main schools of thought: regularisation-based methods, dynamic architectures, and memory-based techniques \cite{wang2024comprehensive, wickramasinghe2023continual}. 

\textbf{Regularisation-based} methods aim to mitigate catastrophic forgetting by adding constraints to the learning process, ensuring that important weights from previous tasks or domains are preserved. EWC \cite {kirkpatrick_overcoming_2017} adds a regularisation term to the loss function to preserve important weights from previous tasks; SI \cite{zenke_continual_2017} uses a similar approach but focuses on the Fisher information matrix to identify important weights; LWF \cite{li_learning_2017} introduces a distillation loss to retain knowledge from previous tasks. \\
\textbf{Dynamic architectures} involve modifying the model's structure to accommodate new tasks while retaining previously learned knowledge. PNNs \cite{rusu_progressive_2022} add new subnetworks for each task while keeping the previously learned parameters frozen; LWTA \cite{srivastava_compete_2013} divides the networks into different blocks and the forward propagation is done only by local winner weights; PackNet \cite{mallya_packnet_2018} prunes the networks and uses masks to filter weights for different tasks; Piggyback \cite{mallya_piggyback_2018} drops the weights training \cite{frankle_lottery_2019} and focuses only on weights masking; a similar approach is used in HAT \cite{serra_overcoming_2018}, but masking the units instead of the weights; SupSup \cite{wortsman_supermasks_2020} simplifies this even more using only a seed to generate random weights and a weighted sum of masks. \\
\textbf{Memory-based} methods utilise stored examples from past tasks to reinforce prior knowledge during training. GDumb \cite{vedaldi_gdumb_2020} randomly selects a subset of exemplars from previous tasks and uses them to train the model; GEM \cite{lopez-paz_gradient_2022} uses episodic memory to store exemplars from previous tasks and ensure that the model does not forget them during training; iCaRL \cite{rebuffi_icarl_2017} utilises stored examples from past tasks to replay and reinforce prior knowledge during training; potentially, one could also use latent generative replay to generate new samples from the previous tasks \cite{pellegrini_latent_2020, graffieti2023generative}.

For a clearer academic evaluation of different CL methods, three main scenarios have been defined and are widely used within the community \cite{van2019three}. In \textbf{Task-incremental learning}, the aim is to incrementally learn a set of distinct tasks based on a given task-id. Whereas in \textbf{Domain-incremental learning}, the context or input distribution varies over time, whilst the task remains constant (e.g. learning to drive in different weather conditions). Finally, in \textbf{Class-incremental learning}, the aim is to incrementally learn to discriminate between a growing number of objects or classes, where task identification is also required. This last scenario is naturally the most difficult to solve, as well as the most applicable within real-world scenarios.

Differently from CL models, Foundation Models (FM) are characterised by their large scale, pre-training on huge amounts of data, and ability to perform a wide range of tasks with a relatively small amount of fine-tuning. These models, such as BERT \cite{devlin_bert_2019}, GPT-3 \cite{brown_language_2020}, and CLIP \cite{radford_learning_2021}, have demonstrated remarkable performance across various benchmarks and applications. However, one of the biggest problems faced by FM models is \textit{homogenisation} \cite{bommasani_opportunities_2022}, where a single unified model trained on diverse data results in generalised internal knowledge representations. While this approach enables model transfer across tasks, it also averages out critical domain-specific nuances, leading to inherent biases from the most dominant data sources. 

More generally speaking, a monolithic AI system, which is a single large model that is trained on a wide range of tasks and domains \cite{zhou2024comprehensive}, is typically pre-trained on massive datasets and then fine-tuned for specific tasks. Such models often require extensive computational resources for training and inference, making them less accessible for smaller organisations and researchers. Additionally, a monolithic model can suffer from knowledge staleness, where the model's performance degrades over time as new data becomes available. Furthermore, their centralised nature poses risks related to single points of failure making them less robust in critical applications. 

On the other hand, continual learning offers a more sustainable and adaptive alternative by enabling models to learn incrementally, adapt to new tasks, and operate efficiently in dynamic settings \cite{wang2024comprehensive}. This paradigm shift is essential for addressing the limitations of monolithic AI and fostering more equitable and resilient AI systems. Continual learning methods could benefit FMs in particular by enabling them to adapt to new tasks and domains without requiring extensive retraining \cite{yang2025recent}, thereby improving their performance over time. By incorporating continual learning strategies, FMs can mitigate issues such as knowledge staleness and inefficiency in adaptation. Furthermore, these methods can help reduce the social and environmental impact of FMs by minimising the need for large-scale retraining, which often requires significant computational resources and energy consumption.

\section{The Need for Continual Learning in the Foundation Model Era}
\label{need-for-CL}

Recently, it has been shown that foundational models such as LLMs have the ability to both ``reason'' and generalise through the use of techniques such as chain of thought (CoT) prompting, by breaking a complex problem into a series of intermediate steps \cite{plaat_reasoning_2024}. This trend of utilising the zero-shot capabilities of LLMs, perhaps with the addition of prompt engineering and later downstream-task fine-tuning, is, however, subject to many different limitations such as the brittleness and inconsistency of the generalised reasoning steps. 

Foundation models such as GPT-3, BERT and DALL-E can quickly become outdated as the real-world data they are trained on changes \cite{brown_language_2020, bommasani_opportunities_2022}. This leads to model staleness over time, where often full model retraining is the only solution used to mitigate this. Recent estimates indicate that the cost of training models of a scale comparable to GPT-4.5 or similar architectures likely reached tens of millions of dollars due to the enormous compute and energy resources required \cite{hoffmann_2022}. Beyond the economic impact, the energy required for such massive compute workloads translates into significant environmental impacts. Studies by Schwartz \textit{et al.} \cite{schwartz_2020} and  Strubell \textit{et al.} \cite{strubell_2020} indicate that the high energy consumption involved in training these models contributes substantially to carbon emissions, emphasising the need for more sustainable approaches.  

State of the art large scale models are typically trained on vast, diverse datasets to capture a wide range of linguistic patterns and knowledge. Whilst this training approach enables impressive generalisation across tasks, it results in a system that is too general to properly address the requirements of individual users or specialised domains \cite{kirk_2023}. In scenarios such as personalised recommendations or adaptive customer support, tailoring interactions based on a user's context or preferences is of paramount importance; however, static models tend to produce generic outputs even after fine-tuning.

Further, conventional personalisation methods, relying on post-training adjustments or Test Time Training (TTT) face significant challenges. In TTT or test time adaptation (TTA), a model adjusts its parameters during inference based on the current input or an auxiliary task, aiming to better align with the data distribution at test time \cite{sun_2019}. Whilst this method allows for on-the-fly adaptation, it frequently requires additional compute post-deployment and may struggle to capture long-term user preferences. Additionally, TTT needs to be combined with continual adaptation with care as it \textit{can} interfere with pre-trained representations if not done correctly, leading to instability and ultimately catastrophic forgetting.

The rising computational and financial demands for training state of the art models have led to a concentration of resources within a small number of organisations. This centralisation means that only a few giants of industry possess the capability to develop, maintain, and update foundation models at scale \cite{bender_2021}. As a result, there is an inherent risk of monopolisation, where control over advanced AI technologies is restricted to those with the resources, potentially limiting progress and diversity within research, raising wider issues around transparency and accountability. For instance, when few entities dominate model training and deployment, issues such as biased data representation and the under-representation of marginalised groups can become more pronounced \cite{pedreschi2024human}. Additionally, the centralised model may raise the barrier to entry for smaller research teams, universities, and independent developers, inadvertently slowing innovation within the AI ecosystem. Furthermore, the monopolisation of AI resources restricts the development of more sustainable and decentralised approaches. 

While the challenges of centralisation, high retraining costs, and limited personalisation have long constrained the evolution of large-scale AI systems, these can each be addressed in different ways by continual learning. There are two fundamentally different types of forgetting in the context of foundational models that necessitate continual learning as a solution: task-shift and time-shift forgetting \cite{shi_continual_2024}. The first, \textit{task-shift forgetting}, arises when a broadly pre-trained model is adapted to a new downstream objective. Without careful safeguards, the updates that confer task skill can overwrite previously acquired general knowledge. This can be effectively mitigated using techniques such as continual pre-training (or domain adaptive pre-training) and continual fine-tuning. The second, \textit{time-shift forgetting}, occurs even when the task remains unchanged: as the external world evolves, the data distribution shifts and a static model's accuracy diminishes unless it is retrained. Additionally, the scalability and modularity of model architectures need to be considered in order to enable models to learn new tasks over time by dynamically composing task-specific modules to solve new tasks \cite{raheja_foundation_2024}. This notion of model compositionality is of the utmost importance to enable models to not only solve novel tasks, but also to be dynamically orchestrated, facilitating interaction with one another in a wider decentralised system of models. Whilst this ecosystem of interactive agents has already begun to come to the fore \cite{shen_hugginggpt_2023, schmidgall_agentrxiv_2025}, all of these models are fundamentally static, requiring centralised retraining over time, which is naturally prohibitive in encouraging more open, democratic and decentralised AI systems. 

\section{Three Key Research Directions for Continual Learning}
\label{research-directions}
\begin{figure}[htbp]
  \centering
  \includegraphics[width=1.0\textwidth]{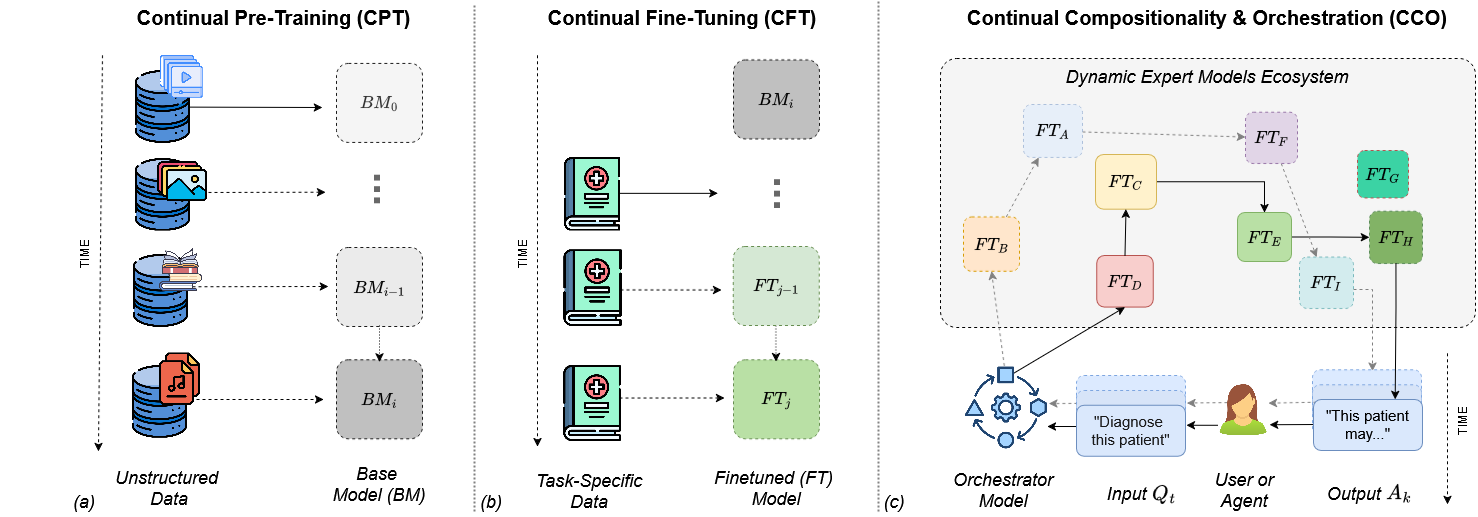}
  \caption{In (a), we see a base model pre-trained on video data is Continually Pre-trained on general corpora spanning different modalities (e.g. audio, images and text). Then in (b), this base model is Continually Fine-tuned over time, resulting in specialised fine-tuning (FT) modules trained on domain-specific datasets, such as medical texts. Finally in (c)---looking at model inference, an orchestrator routes a user’s query through the appropriate FT modules and combines their outputs into a single response. User inputs may change over time, as may the configurations by which models are composed, the processes through which models evolve via Continual Fine-tuning, and the introduction of new models as they become available.}
  \label{fig:benchmark_types}
\end{figure}

Here, we outline the three main research directions that are crucial for the future of Continual Learning, namely Continual Pre-Training, Continual Fine-Tuning and Continual Compositionality \& Orchestration, visualised in Figure \ref{fig:benchmark_types}. In the following sections, we will outline the requirements for each of these separate components, challenges they face, open problems that are yet to be solved and how these methods address the challenges within this context. 

\subsection{Continual Pre-Training}
\label{CPT}
CPT refers to the process of incrementally updating the knowledge of FMs through exposure to new data after their initial pre-training phase \cite{raheja_foundation_2024}. This iterative updating allows FMs to maintain their foundational abilities established during the initial training while simultaneously adapting to assimilate emerging information, thereby extending their operational lifespan and enhancing their adaptability to the ever-changing landscape of data and knowledge \cite{yang2025recent}. 

\subsubsection{Motivation}

We now look to five main motivating factors to advance the field of CPT.

\textbf{Dynamic Knowledge Integration:} CPT is primarily driven by the necessity to keep FMs relevant and effective over time. Even large-scale foundation models can quickly become outdated as data distributions shift and new knowledge emerges \cite{roth_practitioner_2024}.
Static pre-training, performed on a fixed dataset, inevitably leads to models losing touch with rapidly developing fields, where trends, terminologies and societal norms continuously evolve, \textit{e.g.} healthcare, law, technology.
To address this challenge, CPT performs a \textbf{dynamic knowledge integration} by continuously learning on new data streams, enabling FMs to adapt and remain aligned with contemporary information \cite{kuchemann_opportunities_2025}.
Extensive studies have not only demonstrated the necessity of CPT for improved downstream performance, but also shown that when distributional shifts are gradual or somewhat correlated, CPT can effectively help models generalise to previously unseen data \cite{shi_continual_2024}. In particular, CPT enables FMs to handle distributional shifts such as temporal shifts (changes in data over time, leading to model drift), content shifts (changes in topic or domain of the data) and language shifts (introduction of new languages or significant vocabulary changes) \cite{yang2025recent}.

\textbf{Methodological Evolution:} CPT is also motivated by the ongoing improvements in model architectures. Even when there is no new data or major distribution shift, updating foundation models can still be valuable for adopting architectural advances that improve efficiency or performance. For instance, transitioning from an encoder-decoder to a decoder-only architecture benefits from reusing existing pre-trained models, which avoids the need to retrain from scratch and significantly reduces computational cost. CPT enables this kind of update by initialising new models from previous checkpoints, allowing them to retain useful learned knowledge while gradually adapting to architectural changes \cite{parmar_reuse_2024}.

\textbf{Resource Efficiency:} Retraining FMs entirely from scratch on increasingly vast datasets becomes computationally prohibitive over time. CPT significantly reduces these computational burdens by incrementally updating models with new or domain-specific data, circumventing the need for repeated, costly training cycles \cite{shi_continual_2024}. For instance, the LLaMA 4 Behemoth model \cite{Llama4Herd2025}, with its two trillion total parameters, makes full retraining prohibitively expensive, whereas CPT allows for efficient incremental adaptation of FMs.
        
\textbf{Mitigation of Catastrophic Forgetting:} Static models, once trained, are frozen at their initial knowledge cut-off, which creates a gap between initial training and real-world deployment needs. In contrast, CPT fosters a continual evolution of models, bridging the gap between initial pre-training and practical, lifelong learning scenarios. Notably, recent studies find that CPT can make models more robust to catastrophic forgetting of earlier knowledge, especially when using self-supervised objectives, highlighting CPT as a key enabler for foundation models to learn continuously like humans \cite{cossu_continual_2024}.

\textbf{Selective Forgetting:}
In addition to mitigating catastrophic forgetting, CPT has the potential to support selective forgetting, where specific information embedded in a foundation model is intentionally removed over time \cite{golatkar_eternal_2020}. This capability is particularly relevant as large-scale models may inadvertently memorise sensitive, outdated, or harmful content during pre-training \cite{wang_selective_2025}. Recent work has emphasised the importance of enabling continual forgetting to remove such undesirable knowledge while preserving overall model performance \cite{zhao_continual_2024}. While earlier approaches explored this problem through fine-tuning, recent efforts have extended it into the CPT phase. For example,  Zhu \textit{et al.} introduce a regularised CPT method that enables the removal of backdoor behaviours in language models while maintaining their functionality on clean data \cite{zhu_removing_2023}. These developments highlight selective forgetting within CPT as a promising direction for maintaining the safety, privacy, and reliability of foundation models over time.

\subsubsection{Challenges \& Open Problems}
CPT is still in the early stages of development, and bridging the gap between research and production remains challenging: while CPT techniques show promise in controlled experiments, their long-term stability and effectiveness over months of deployment in real-world settings remain under-explored \cite{shi_continual_2024}. This highlights several key challenges that must be addressed to make CPT viable in practice, including:

\textbf{Handling catastrophic forgetting:} Catastrophic forgetting, the phenomenon where a continually updated model loses previously acquired knowledge, remains a critical challenge in CPT \cite{brinner_enhancing_2025, li_examining_2024}. Although scaling up pre‑training tends to enhance knowledge transfer and resilience against forgetting during downstream CL, excessively extensive pre‑training can significantly increase the risk of forgetting \cite{mehta_empirical_2023,wu_continual_2024}. Recent studies indicate that self-supervised CL exhibits significantly reduced catastrophic forgetting compared to supervised approaches \cite{zhu_ctp_2023, cossu_continual_2024}. Indeed, self-supervised pre-training has the advantage of reduced forgetting during upstream tasks, though effectively balancing upstream CL with downstream continual adaptation remains an open research question \cite{zhang_slca_2023}. Additionally, model scale plays a pivotal role: larger models consistently demonstrate lower perplexity (an indicator of how unfamiliar or novel a document is to a language model) and reduced forgetting, whereas smaller models, despite achieving substantial learning gains, tend to exhibit the most pronounced forgetting effects \cite{yildiz_investigating_2024}.

\textbf{Balancing Efficiency vs. Model Drift:} FMs often have hundreds of billions of parameters, so retraining them on every new dataset is extremely computationally expensive \cite{raheja_foundation_2024}. Practical CPT must therefore be computationally efficient, for instance by updating only a subset of parameters or using limited data, but this can exacerbate the stability–plasticity dilemma. Insufficient or biased updates may lead to model drift, where performance on original domains degrades or the model’s behaviour shifts unpredictably. Empirically, a “stability gap” has been observed: when an LLMs is first continually pre-trained on a new domain, its performance drops initially (due to distribution shift) before recovering \cite{guo_efficient_2024}. Balancing efficient adaptation with stability (avoiding regressions on prior knowledge) is an open problem.
         
\textbf{Avoiding reinforcement of biases in pre-training:} Continuously ingesting new data can also reinforce biases or fairness issues if not carefully controlled. If the incoming data is skewed or uncurated, the model might amplify these biases over time, undermining responsible use. For example, biomedical FMs trained on federated data must address fairness across diverse populations while preserving privacy \cite{li_open_2025}. Ensuring that continual updates do not degrade the model’s ethical alignment (e.g. with respect to bias and fairness) is crucial \cite{raheja_foundation_2024}. Methods to detect and mitigate bias drift during CPT (and to curate update data) are largely lacking and represent an important research frontier.

\subsubsection{Potential Solutions \& Future Research Directions}
To address the challenges of CPT and unlock its full potential, several promising solutions and future research directions have been proposed, including:

\textbf{Incremental pre-training strategies:} Recent works propose reusing or initialising from previous model weights to maintain continuity, as in recyclable tuning methods that carry over knowledge from an old model to a new one \cite{yang2025recent}. Recent work has demonstrated the benefits of structured or multi-stage CPT. For instance, in \cite{mendieta_towards_2023}, a two-step CPT was shown to enable a mixed-language neural machine translation system (first adapting to a language domain, then to a specific translation task) effectively. In \cite{dalla_sequential_2024}, Dalla Noce \textit{et al.} introduce a sequential CPT framework for neural machine translation, where a model is progressively exposed to new language pairs or domains in multiple stages. Their study finds that incrementally adding new languages during the pre-training phase does not substantially degrade the model’s performance on previously seen language pairs during fine-tuning. Furthermore, incorporating CL strategies such as data rehearsal can further reduce performance loss on earlier language pairs compared to purely incremental pre-training but comes with increased computational cost during the training phase. This highlights a practical trade-off between training efficiency and performance robustness: reasonable downstream performance can be achieved through incremental pre-training, but further performance improvements can be attained when accepting the additional computational cost associated with CL strategies.

\textbf{Selective memory and rehearsal methods:} To combat forgetting, memory replay methods have shown promise in the context of foundation models. Rather than relying only on the latest data, the model can intermittently rehearse on representative samples of past data (or tasks). In practice, storing raw past data for a foundation model may be impractical or raise privacy concerns such that recent work leverages latent replay, where the model saves a cache of compact feature representations or embeddings of past examples instead of the raw inputs \cite{raheja_foundation_2024}. During CPT, these stored latent vectors can be replayed through the model to reinforce previously learned concepts. This memory-efficient replay has been shown to significantly mitigate forgetting in vision models and is especially valuable when sharing raw data is prohibited (e.g. user data privacy). Besides replay, selective sampling strategies can be used rather than naively mixing new data; the training scheduler might interleave the most informative or relevant examples carefully chosen from older tasks or emphasise difficult examples that the model is starting to forget. There is evidence that the order and composition of training data in CPT can greatly affect retention \cite{roth_practitioner_2024}. For example, Xie \textit{et al.} use perplexity and embedding similarity metrics to select a fraction of a domain corpus that achieves comparable adaptation with far less training cost \cite{xie_efficient_2024}. Such sampling not only improves efficiency but can also prevent the model from drifting too far by ensuring the new training distribution is aligned with the model’s original knowledge. Additionally, some approaches intermix new data with a small portion of the original pre-training data (or a similar distribution) during updates, explicitly to reduce distribution shift \cite{guo_efficient_2024}. This kind of rehearsal or data mixing has been shown to narrow the stability gap and avoid performance deterioration on earlier tasks. Going forward, developing principled sampling and replay policies (potentially guided by uncertainty, importance weighting, or task identities) is a key research direction to make CPT robust and scalable.

\textbf{Self-supervised continual adaptation techniques:} CPT largely relies on self-supervised learning objectives such as predicting masked tokens, next sentence prediction, and image-text contrastive learning, because these allow the use of unlabelled streaming data. An intriguing finding from recent research is that self-supervised objectives can themselves be leveraged to improve CL. In \cite{cossu_continual_2024}, Cossu \textit{et al.} provide strong empirical evidence that continuing pre-training models in a self-supervised manner yields better knowledge retention than supervised training in the continual setting. Intuitively, self-supervised learning updates may be softer or more diffuse in the parameter space (since they capture broad patterns in data) compared to task-specific fine-tuning which might overwrite more specialised parts of the model. In \cite{ostapenko_continual_2022}, Ostapenko \textit{et al.} also observed that models pre-trained with broader or more diverse self-supervised signals tend to forget less and transfer better in downstream sequential tasks. This suggests that self-supervised CPT is a promising avenue: as new unlabelled data comes in, one can design auxiliary objectives that encourage the model to integrate new information while maintaining consistency with prior representations. Techniques like contrastive learning on a replay buffer, or predictive modelling that ties new concepts to old ones, fall in this category. Moreover, self-supervision can be combined with light supervision or prompts in an autonomous CL setup (where a model might generate pseudo-labels or questions on new data and learn from them). Overall, self-supervised learning-based CPT not only provides a means to utilise vast unlabelled streams, but also appears to inherently mitigate forgetting, making it a key research direction for lifelong FMs.

\subsection{Continual Fine-tuning}
\label{continual-fine-tuning}

Continual fine-tuning is the practice of applying a stream of lightweight, task‑specific updates to a model after deployment, allowing it to evolve alongside newly arriving data rather than remaining fixed after a single adaptation.

\subsubsection{Motivation}

Fine-tuning is attractive because it it far cheaper than full retraining, requires only task-specific data, and can be executed on modest hardware \cite{houlsby2019parameterefficienttransferlearningnlp, hu2021loralowrankadaptationlarge}. Yet, once the weights are written to disk, the model is frozen again. In realistic deployments, data arrives as a stream, such as customer queries, sensor logs, freshly published documents --- therefore the ability to fine-tune \textbf{continually} is crucial \cite{aggarwal2024exploringcontinualfinetuningenhancing, beaulieu2020learningcontinuallylearn}. CFT turns a one-shot adaptation step into a standing capability that (i) personalises responses for each user or organisation \cite{zhang2025cloracontinuallowrankadaptation}, (ii) keeps proprietary data on-premise for privacy compliance \cite{fan2023fatellmindustrialgradefederated}, (iii) reacts quickly to domain drift without the latency of retrieval augmented generation (RAG) pipelines or very long context windows \cite{team2024gemini}, (iv) does all of this with a fraction of the compute budget needed for repeated full-scale updates \cite{liu2024fisherinformationbasedefficientcurriculum, soutif2023comprehensive}. The need for CFT, the process of incrementally fine-tuning a model to help it adapt to downstream tasks that involve shifting data distributions and temporal changes \cite{aggarwal2024exploringcontinualfinetuningenhancing}, cannot be overstated.

Although FMs have demonstrated impressive versatility across different tasks, with the ability to generalise effectively to various domains, their static nature limits the capacity to incorporate new knowledge, adapt to specialised fields, and personalise outputs over time. CFT presents an important opportunity to make foundation models more flexible, efficient and responsive to real-world changes, making them more useful in more dynamic environments. While continual pre-training focuses on updating a model's general representations using broad, often unlabelled data, continual fine-tuning instead aims to incrementally adapt the model to specific downstream tasks using labelled or structured data, with an emphasis on retaining prior knowledge while learning new information. 

\subsubsection{Challenges \& Open Problems}

CFT in the context of foundation models like LLMs comes with several challenges such as:

\textbf{Balancing Specificity vs. Generalisation}: CFT must maintain a delicate equilibrium between tailoring a model to a specific downstream task and preserving its broad, generalisable knowledge. When a model undergoes CFT, its internal representation becomes optimised to capture patterns necessary for solving a specific downstream task. While this adaptation enhances the performance on domain specific tasks, it risks eroding the broad, general-purpose representations learned during pre-training. This comes back to the stability-plasticity dilemma where models must remain plastic enough to integrate task-specific knowledge while being stable enough to retain the broad representations acquired from prior experiences \cite{10.1016/j.neunet.2019.01.012}.

\textbf{Efficient Adaptation Without Catastrophic Forgetting}: In a similar vein to CPT, catastrophic forgetting is also experienced by CFT methods, however within this context \textit{efficient adaptation} also needs to be considered. This concept refers to the process of updating a pre-trained foundation model to perform well on new tasks, domains, or data distributions while minimising computational resources, data requirements, and training time. In the context of LLMs and other foundation models, efficiency has become increasingly critical as these models grow to billions of parameters. Efficient finetuning techniques like LoRA \cite{hu2021loralowrankadaptationlarge} allow large pre-trained models to be adapted to downstream tasks by updating only a fraction of the model's original parameters, but these techniques are still prone to catastrophic forgetting.

\textbf{Data efficiency and privacy concerns in continual fine-tuning}: CFT deals with the dual challenges of data scarcity and privacy concerns, especially in specialised domains to effectively adapt models to new tasks. As foundation models are adapted to increasingly specialised domains, high-quality, domain-specific data becomes progressively scarcer \cite{du2025privacyfinetuninglargelanguage}.

\subsubsection{Potential Solutions \& Future Research Directions}

Despite these limitations, numerous methods have been developed to tackle these challenges. 

\textbf{Parameter Efficient Fine-Tuning}
A key methodological tool in the context of CFT is represented by Parameter Efficient Fine-Tuning (PEFT) methods.
These techniques aim to achieve performance comparable to or even surpass full model fine-tuning while updating only a small number of trainable parameters, either by selectively updating a subset of the model's parameters \cite{hu2021loralowrankadaptationlarge} or introducing new task-specific parameters \cite{houlsby2019parameterefficienttransferlearningnlp}.
PEFT methods are particularly advantageous in continual learning scenarios, where models must adapt to a sequence of tasks without forgetting previously learned information.

By updating only a limited number of parameters, PEFT approaches reduce computational overhead and, when applied properly, can help mitigate the risk of catastrophic forgetting. Among these PEFTs, Prompt Adapters and LoRA are the most widely used. LoRA works by introducing low-rank updates to the pre-trained model weights, expressed as:
    \begin{equation}
    W = W_0 + BA
    \end{equation}
    \noindent
where the pre-trained model $W_0$ is kept frozen, while the low-rank matrices $A$ and $B$ are updated.

Prompt-based techniques like L2P and CoDA Prompt \cite{wang2022learningpromptcontinuallearning,smith2023codapromptcontinualdecomposedattentionbased} incrementally learn from novel data by designing task-specific prompts that guide the model's attention toward relevant information for each new task, facilitating seamless integration of new knowledge without overwriting existing capabilities.

Similarly, LoRA-based CL approaches, such as C-LoRA and DualLoRA \cite{zhang2025cloracontinuallowrankadaptation,chen2024duallowrankadaptationcontinual}, enhance LoRA's applicability in CL by introducing mechanisms like learnable routing matrices and orthogonal subspaces to manage parameter updates across tasks, thereby reducing computational overhead and mitigating catastrophic forgetting.

Adapter techniques like Continuous Adapter (C-ADA) and Adapter-based Continual Learning (ACL) \cite{zhang2023adapterlearningpretrainedfeature,gao2024promptlearningcontinualadapter} instead offer more efficient solutions for CL. C-ADA introduces a Continual Adapter Layer that extends weights for new tasks while freezing old ones, preserving prior knowledge. It also employs a Scaling \& Shifting module to align feature spaces between pre-training and downstream datasets. Similarly, ACL utilises lightweight, task-specific adapters within a fixed pretrained feature extractor and incorporates a task-specific head that groups previously learned classes into an "out-of-distribution" category, facilitating effective feature discrimination.

\textbf{Model Merging}: A particularly valuable approach when facing dynamic environments is given by Model Merging.
The key point in this class of methods is to combine multiple specialised models learned over time, to create systems that preserve knowledge while adapting to new tasks.
The central challenge in model merging is parameter interference, where integrating different models leads to performance degradation. Recent research has developed several innovative solutions to this problem. TIES-MERGING \cite{yadav2023tiesmergingresolvinginterferencemerging} addresses interference by strategically resetting minimally changed parameters and resolving sign conflicts between models. In contrast, DARE \cite{yu2024languagemodelssupermario} employs a different strategy by randomly dropping redundant delta parameters and rescaling the remaining ones, effectively sparsifying merged models without significant performance loss.

While early model merging techniques focused on the static combination of pre-existing expert models, more recent approaches support dynamic integration as new tasks emerge over time. MagMax \cite{marczak2024magmaxleveragingmodelmerging} introduces sequential fine-tuning with maximum magnitude weight selection to effectively incorporate new information while preserving earlier learning. Representation Surgery \cite{yang2024representationsurgerymultitaskmodel} tackles representation bias by inserting lightweight task-specific modules that realign internal representations between merged models. Adaptive LoRA Merging \cite{coleman2024adaptive} moves beyond fixed-weight combinations by dynamically computing merging coefficients that balance contributions from new and old domains.

Recent trends in CFT have shifted towards the adaptive integration of lightweight modules, like adapters \cite{houlsby2019parameterefficienttransferlearningnlp} and LoRA \cite{hu2021loralowrankadaptationlarge}, in dynamic environments. This shift enables seamless integration of new tasks without extensive retraining of large models. By merging these modular components on demand, systems can efficiently handle real-world challenges while remaining practical for large-scale deployment.

\textbf{Meta Learning for Continual Adaptation}: An alternative perspective on the challenge of CFT is offered by meta learning approaches for continual adaptation.
In fact, these methods integrate adaptability into the core learning objective, enabling models to rapidly adjust to new tasks with minimal data and computation. Traditional meta-learning approaches like Model-Agnostic Meta-Learning (MAML) \cite{finn2017modelagnosticmetalearningfastadaptation} operate by finding parameter initialisations that enable rapid adaptation across a distribution of tasks. When applied to CL scenarios, these methods can be extended to discover parameter configurations that not only adapt quickly but also resist catastrophic forgetting. For instance, ANML \cite{beaulieu2020learningcontinuallylearn} uses a neuromodulatory network that enables the model to focus on relevant tasks while minimising interference from previously learned tasks.

Recent works have combined meta-learning with parameter-efficient fine-tuning techniques to enhance CL. AutoLoRA \cite{zhang2024autoloraautomaticallytuningmatrix} introduces a meta-learning framework that automatically identifies the optimal rank for each LoRA layer, improving adaptation efficiency to new tasks while maintaining performance on previous ones. Similarly, Meta-LoRA \cite{li-etal-2025-meta} presents a memory-efficient method for automatic sample re-weighting during fine-tuning, facilitating efficient continual adaptation across various domains. These approaches exemplify the potential of meta-learning to enhance the adaptability and efficiency of foundation models in dynamic environments.

\textbf{Federated Learning (FL) and Decentralised Fine-Tuning Strategies}: Moving towards the direction of a distributed and decentralised AI development, FL is essential for effectively adapting large FMs across organisations, while preserving data privay and optimising computational resources.
When applied to FMs, however, FL faces unique challenges. The vast number of parameters in modern FMs makes server-client communication prohibitively expensive. Different clients naturally generate data with varying distributions, creating potential conflicts in optimisation objectives. Frameworks like FATE-LLM \cite{fan2023fatellmindustrialgradefederated} enable collaborative training of LLMs by employing parameter-efficient fine-tuning methods and incorporating privacy-preserving mechanisms. FibecFed \cite{liu2024fisherinformationbasedefficientcurriculum} enhances this approach by utilising Fisher information for adaptive data sampling and dynamically selecting layers for global aggregation, thereby improving both performance and fine-tuning speed. Additionally, FedRewind \cite{palazzo2024fedrewindrewindingcontinualmodel} introduces a decentralised model exchange strategy inspired by continual learning principles, addressing data distribution shifts and enhancing generalisation performance in federated settings.

\subsection{Continual Compositionality \& Orchestration}
\label{continual-compositionality}

Continual Compositionality \& Orchestration refers to the dynamic integration of multiple AI agents over time, to solve higher-level tasks.
It is the key component towards a distributed and decentralised AI framework.

\subsubsection{Motivation}
Large models solved most of the tasks addressed by AI methods for decades.
In fact, using a Transformer-based architecture, pre-trained on some large dataset, we now know how to solve almost every task, provided enough data and computational power.
Broadly speaking, modern FMs have achieved super-human performance on most of the traditional machine learning benchmarks, making them obsolete and less relevant for current AI research.

For this reason, we are now shifting towards higher level tasks, which require a higher level of intelligence, that current \textit{state-of-the-art} models do not exhibit.
This trend is demonstrated by the emergence of several more complex benchmarks, that could drive the research beyond current LLMs.
One such benchmark is ARC-AGI \cite{chollet2019github}, which consists of simple grid transformations.
Despite being simple for a human solver, it poses great challenges, requiring strong abstraction skills and inductive reasoning.
Another example is BIG-bench \cite{srivastava2022beyond}, a collection of more than 200 tasks, designed to test the limits of current large models.
The benchmark covers a broad set of tasks, including linguistics, mathematics, common-sense reasoning, social bias detection and more.
In both cases, human performance significantly surpasses the current best models, highlighting the need for alternative solutions.

Furthermore, latest large models, \textit{e.g.} GPT 4.5, demonstrate that we are rapidly moving towards a diminishing returns regime, meaning that just increasing the model dimensions and the number of GPUs employed is no longer enough to obtain noticeable improvements \cite{luo2024has}.
We believe that a paradigmatic change is needed in the AI community in order to push research forward and to obtain more intelligent behaviours.

Our proposal is to address some of the shortcomings that current solutions exhibit, particularly regarding scalability and sustainability matters, with \textbf{Continual Compositionality and Orchestration} approaches.
Nowadays, the \textit{de facto} standard is to have single, monolithic models, trained once and deployed without any guarantees on their utility over time.
CCO instead represents a framework built on the communication between multiple AI models, which share their knowledge and skills in terms of model parameters, deep representations or final predictions.

Within this paradigm, the orchestration among the \textit{agents} becomes crucial: depending on the task, different modules are selected to be composed in various fashions.
In other words, rather than adjusting a single network to a dynamic scenario, CCO employs a modular approach, where different module compositions can be used to adapt to non-stationary environments over time.
Also, such a framework does not impose constraints on the scalability of the system, \textit{i.e.} the number of modules involved, and it is more sustainable \textit{by design}, since the modules are trained once and then re-used over time in multiple ways.

\subsubsection{Challenges \& Open Problems}
With the advancements in capabilities of LLMs, there has been a growing research focus on building LLM-based agent architectures, in which multiple models are composed and coordinated to solve complex tasks \cite{tran2025multi,wang2024survey}. We can place agents on a spectrum based on their level of autonomy in order to differentiate their architectures. On one end, fully autonomous systems, in which agents interact with significant freedom; on the other, predictable and structured workflows, in which agents follow  predefined steps and communication patterns \cite{liu2025advanceschallengesfoundationagents}. More structured and predetermined workflows might be preferable for domains that require more precision and accountability, such as mathematical reasoning, scientific research, law, medicine, and software development.

\textbf{Task Decomposition and Specialisation}:
An effective strategy to enable LLMs to solve complex problems is to break them down into simpler, more manageable sub-tasks \cite{wei2022chain,xi2025rise,raheja_foundation_2024}. However, fully automating the planning and decomposition of tasks into sub-tasks with LLMs is an area of open research. Some studies propose decomposing a problem with a single LLM request that generates a series of sub-steps \cite{ichter2023can,xu2023rewoo, raman2022planning}. Other studies propose more advanced search-based approaches, which iterate and further decompose each sub-task into smaller steps when necessary; the final execution plan can then be organised into a tree-like structure \cite{yao2023tree}. One major challenge of this area of research is generating plans for domain-specific problems; the use of external planners is one of the solutions that have been proposed to address this issue \cite{wang2024survey}. Additionally, the planning abilities of LLMs might still be limited by their lack of human-level comprehension of world dynamics and the ability to apply causal reasoning to them \cite{liu2025advanceschallengesfoundationagents}.

Within this context, Mixture-of-Experts (MoE) models have emerged as a prominent line of research, offering a natural implementation of the \textit{divide and conquer} paradigm.
MoE architectures aim to scale model capacity efficiently by activating only a subset of experts per input, leading to improved performance \cite{du2022glam, lin2024moe}.
However, a core challenge lies in achieving effective \textit{expert specialisation} --- ensuring that each expert acquires unique and distinct skills, with little overlap with the others.
Preliminary works suggests that expert models specialise on superficial patterns, such as token IDs, rather than extracting high-level semantic information \cite{xue2024openmoe}.
Although recent efforts have proposed solutions towards more meaningful expert specialisation \cite{dai2024deepseekmoe}, we still lack a clear understanding of these mechanisms, especially when considering distributed and decentralised AI frameworks.
Further research is needed to ensure that MoE models can robustly and adaptively decompose complex tasks in dynamic, multi-agent environments.

\textbf{Role-based Collaboration and Interactions}:
In a compositional framework, interactions among agents represents a key ingredient.
One of the most common strategies to compose multiple agents to work together in solving a task is the so-called role-based collaboration. LLM-based agents assume clearly defined, specialised roles (such as domain experts, assistants, etc.) in order to solve a higher level goal, with each of them being assigned individual sub-tasks by other agents. Optimal role assignments and agent adaptability to dynamic tasks requirements are still areas of open research \cite{tran2025multi}. One challenge is that, while LLMs are able to simulate many common roles, there are still many roles that they might not be able to capture accurately, such as uncommon roles rarely seen in the training corpus, or roles corresponding to human characters with particular cognitive-psychological traits \cite{fischer2023reflective,li2023emotionprompt,wang2024survey}.

Additionally, agent interaction play an essential role for multi-round tasks, where multiple iterative feedbacks loops --- both from the environment and from other agents --- are required to achieve the objective.
Such tasks demand dynamic coordination, contextual adaptation, and the ability to reason over partial progress.
Recent studies have begun to explore this promising topic, introducing novel solutions for LLM collaboration \cite{wangsegllm, zhou2025collaborative}.
This underscores the relevance of multi-agent interactions as a key challenge for CCO.

\textbf{Propagation of errors}:
Lastly, the propagation of errors is an additional open problem in ensuring robustness in model integration. Erroneous outputs, hallucinations and biases from one agent can have cascading effects, getting amplified and spread through model interactions and impacting the whole system \cite{tran2025multi}.

\subsubsection{Potential Solutions \& Future Research Directions}
The concept of multiple models collaborating within a shared environment is well established in the AI field, and is rooted in the foundational definition of \textbf{multi-agent systems (MAS)}.
Such systems consist of multiple autonomous agents, each with its own goals and motivations, that are capable of interacting with one another.
For this reason, a central focus of this paradigm is on \textbf{cooperation} and \textbf{coordination} among agents.\\
The CCO framework can be viewed as a concrete instantiation of MAS, where models such as LLMs must be orchestrated to achieve a common objective.
A key research direction in this context is the dynamic selection and composition of the most appropriate agents for a given task.
Indeed, in the CCO framework the goal is to enable automatic models composition, that can evolve dynamically over time in response to variations in the environment or in the task given by the user.
This marks a significant advancement over existing compositionality frameworks, \textit{e.g.} LangGraph\footnote{https://www.langchain.com/langgraph}, AutoGen \cite{wu2023autogen}, in which the orchestration is largely static and predefined by the programmer, thereby limiting both flexibility and the capacity to generalise across tasks and environments.

Another component enabling continual learning capabilities in the CCO framework could be the use of experience accumulation modules, in the form of memory modules and skill libraries, as proposed in works like GITM, Voyager, AppAgent and MemPrompt \cite{wang2024voyager,zhu2023ghostminecraftgenerallycapable, zhang2023appagent, madaan2022memory}. These approaches allow models to dynamically acquire new knowledge and skills, as a result of interactions with other agents, humans and the environment \cite{wang2024survey,xi2025rise,sumers2024cognitive}. Such knowledge and skills can be stored in natural language or code form, and later retrieved and added to the model input context as needed. These mechanisms rely on test-time inference and incorporate the new knowledge and skills in the input context window of the model. One limitation might be that, depending on the LLM used, the context window capacity might limit the amount of task information that can be incorporated; however, research advancements in this area are enabling ultra-long context windows of 1M tokens or more \cite{team2024gemini}, albeit at the expense of added compute. 

An additional crucial aspect lies in the communication between agents.
In the case of LLM agents, communication can occur through natural language, which has the added benefit of being easily interpretable by human users --- a property that enhances transparency and human-in-the-loop control \cite{li2023camel}.
However, the CCO framework is designed to be model-agnostic and general-purpose, extending beyond language models to integrate a diverse set of AI components --- such as computer vision models, time series processors, rule-based systems, symbolic modules or even hard-coded functions.\\
To support such heterogeneity, the system requires a robust communication protocol, that accommodates decentralisation, asynchrony and different data formats, while enabling efficient knowledge exchange.

Furthermore, an important open question in the design of the CCO protocol is in what type of knowledge should be shared among agents, to maximise collaboration without unnecessary overheads.
Depending on the use case, this could include (i) model parameters, either in entirety or specific subnetworks / modules; (ii) the model internal representation, \textit{e.g.} latent vectors or output logits; (iii) training data, as raw samples or abstracted via data generators.

\section{The Future of Continual Learning: From Niche Research to AI’s Next Paradigm}

Over the past decade, Continual Learning (CL) has established itself as a prominent research area exploring a variety of domains, problem settings and applications \cite{wang2024comprehensive}. In Computer Vision, significant effort has been committed to \textit{class incremental} and \textit{domain incremental} scenarios, where models must progressively recognise new categories without forgetting previously learned ones, even as the input domains evolve \cite{liu2023incremental}. In Reinforcement Learning, CL has focused on the ability of agents to adapt dynamically to evolving environments while retaining past knowledge and abilities, particularly in \textit{task incremental} and \textit{multi-task} settings \cite{khetarpal2022towards}. More recently, CL research has extended its scope to include LLMs and Foundational Models, where the challenge lies in enabling models to continuously acquire new linguistic capabilities or domain knowledge without catastrophic forgetting \cite{raheja_foundation_2024,shi_continual_2024}. Another critical aspect is memory management in lifelong learning AI systems. Several studies examine the inherent trade-off between limited computational and storage resources and the ever-growing volume of data that CL models are expected to handle. Researchers have proposed a wide range of methods to strike a balance between learning efficiency and memory constraints such as dynamic memory buffers, experience replay mechanisms, architectural approaches, and regularisation techniques \cite{wang2024comprehensive,wickramasinghe2023continual}. Leveraging this attention from the community, CL has become a well-established and recognised field, providing both theoretical foundations and practical methodologies for building adaptive, robust, and memory-efficient AI systems.

Given the increasingly consolidated position of CL within the research community, we believe the time is right for the field to take a decisive step forward. Rather than representing a separate area of study, CL should become a critical and fundamental component in the prototyping and evolution of modern AI systems. In particular, the rapid rise of foundational models has captured the attention of both academia and industry, asserting itself as one of the most prominent and transformative trends in contemporary AI research. These models showcase broad generalisation capabilities across tasks and modalities; however, they still exhibit a critical limitation: their knowledge is inherently static and fixed at training time. This immutability poses a major challenge in dynamic real-world environments, where new data and information continuously emerge.

To address this gap, foundational models must evolve towards true continuous adaptation, progressively updating, refining, and extending their knowledge over time. Continual learning principles can provide a concrete and resourceful solution in this context. As previously discussed, \textbf{Continual Pre-training} and \textbf{Continual Fine-tuning} represent emerging research directions that aim to integrate the principles of CL with large-scale models, enabling them to remain updated, accurate, and contextually relevant, thereby mitigating outdated or incorrect outputs \cite{yang2025recent}. However, CPT and CFT inherently require relatively low-frequency adaptation cycles and depend on substantial computational resources. Thus, while beneficial, these approaches alone might not address the dynamic adaptability required by real-world applications fully. 

Real-world change, in contrast, is often expressed at the level of \textit{orchestration}: new tools appear \cite{shen_hugginggpt_2023}, regulations shift, a CoT must be revised, or a group of agents must re-organise to solve an emergent sub-problem \cite{schmidgall_agentrxiv_2025}. This layer is inherently \textit{high-frequency and pervasive}, with potential updates required minutes or even seconds after new data arrives, making repeated offline training cycles impractical \cite{kumar_2025}. Here \textbf{Continual Compositionality and Orchestration} is not merely advantageous - it is indispensable. CCO treats an AI system as a living assembly of modules composed of prompt routers, domain experts, external tools, episodic memories --- all of which can be composed and adapted on the fly. Continual learning supplies the two capabilities such a system requires: \textit{rapid adaptation} to integrate the next tool or component and \textit{memory consolidation} to stabilise useful compositions so that they can be re-used rather than rediscovered.

Foundational models are generally effective across broad tasks; however, when maximising performance in specialised areas, such as mathematics or physics, monolithic models may underperform compared to models specifically distilled or trained for those individual tasks \cite{gou2021knowledge,yang2024survey}. We believe that creating and deploying CL models which can continually evolve and combine knowledge from different sources represents a promising and sustainable architectural solution for AI systems. Moreover, developing models that are a mixture of many components \cite{raheja_foundation_2024, carta2022ex} can allow for the distribution of computation and decentralisation of AI systems. Different institutions can collaborate and contribute to create the ad-hoc models that synergise to achieve the final and complete AI system.

Furthermore, aligning with recent advances, foundational models can also be improved through human feedback within a Reinforcement Learning from Human Feedback (RLHF) framework, where humans actively guide model evolution, shaping future AI outputs and capabilities in turn \cite{pedreschi2024human}. This human-AI feedback loop can help to ensure that AI development is in line with human preferences and can also be applied within the context of multi-agent systems. Human-in-the-loop approaches in agentic LLM systems can be used to provide guidance, supervision and feedback to individual agents, facilitating alignment with human preferences \cite{xi2025rise,kenton2021alignment,du2022read,cai2023human}. 

In summary, Continual Learning, specifically continual compositionality and orchestration, represents not only a promising research direction, but the cornerstone of AI's next paradigm shift. By transitioning from incremental improvements within individual models to dynamically composable and collaborative AI ecosystems, CL can drive a new generation of adaptive, scalable and human-aligned AI systems. 

\section{Conclusion}
\label{conclusion}

The remarkable capabilities of large foundational models highlight their potential in solving complex, diverse tasks across multiple domains. However, despite their robust generalisation abilities, these models are inherently static and struggle to adapt continually to evolving real-world data and tasks. To overcome these limitations, continual learning emerges as an indispensable tool, offering a multitude of methodologies to enhance adaptability, efficiency and sustainability within foundational models. 

In this paper, we have highlighted three pivotal areas of continual learning critical to the evolution of FMs: continual pre-training, continual fine-tuning, and continual compositionality and orchestration. CPT equips very large, organisation-scale models with the mechanisms to incrementally incorporate new knowledge, capabilities or methodologies, maintaining their relevance and mitigating catastrophic forgetting through adaptive, resource-efficient updates. As such, it is mainly within the remit of industrial laboratories and cloud providers who possess the necessary data and compute. CFT remains valuable, although comparatively secondary, to enable precise adaptation to specialised tasks and domains, effectively balancing specific task performance with the retention of generalisable knowledge. Techniques such as PEFT, meta-learning and model merging were identified as promising approaches to achieving effective adaptation while managing computational resources and limiting data drift.

CCO, by contrast, is where academic research can and should place its primary emphasis. Moving from monolithic models to a decentralised ecosystems composed of specialised, modular agents facilitates adaptability, enhances scalability and reduces centralisation risks by enabling modular model replacement, upgrade and collaborative interaction. By studying and advancing CCO, the research community can catalyse an open, decentralised, and circular economy of AI components. Such a decentralised ecosystem not only encourages innovation and democratises access, but also mitigates the computational and environmental costs associated with continually retraining large, static models. 

Ultimately, continual learning positions itself not merely as an optional enhancement but as a foundational requirement for future AI systems. As AI evolves from static to dynamic, from centralised to decentralised, and from monolithic to modular, the integration of continual learning methodologies will be crucial. Embracing continual learning will therefore be instrumental in building resilient, flexible and context-aware AI systems, capable of sustainably adapting to the ever-changing landscape of real-world challenges.

\section*{Acknowledgements}

Research partly funded by PNRR - M4C2 - Investimento 1.3, Partenariato Esteso PE00000013 - ”FAIR - Future Artificial Intelligence Research” - Spoke 1 ”Human-centered AI”, funded by the European Commission under the NextGeneration EU programme.

\bibliography{Bibliography}

\begin{thebibliography}{134}
\expandafter\ifx\csname natexlab\endcsname\relax\def\natexlab#1{#1}\fi
\providecommand{\url}[1]{\texttt{#1}}
\providecommand{\href}[2]{#2}
\providecommand{\path}[1]{#1}
\providecommand{\DOIprefix}{doi:}
\providecommand{\ArXivprefix}{arXiv:}
\providecommand{\URLprefix}{URL: }
\providecommand{\Pubmedprefix}{pmid:}
\providecommand{\doi}[1]{\href{http://dx.doi.org/#1}{\path{#1}}}
\providecommand{\Pubmed}[1]{\href{pmid:#1}{\path{#1}}}
\providecommand{\bibinfo}[2]{#2}
\ifx\xfnm\relax \def\xfnm[#1]{\unskip,\space#1}\fi
\bibitem[{French(1999)}]{french_catastrophic_1999}
\bibinfo{author}{R.~M. French},
\newblock \bibinfo{title}{Catastrophic forgetting in connectionist networks},
\newblock \bibinfo{journal}{Trends in Cognitive Sciences} \bibinfo{volume}{3} (\bibinfo{year}{1999}) \bibinfo{pages}{128--135}. \URLprefix \url{https://www.cell.com/trends/cognitive-sciences/abstract/S1364-6613(99)01294-2}. \DOIprefix\doi{10.1016/S1364-6613(99)01294-2}, \bibinfo{note}{publisher: Elsevier}.
\bibitem[{Ditzler et~al.(2015)Ditzler, Roveri, Alippi, and Polikar}]{ditzler_2015}
\bibinfo{author}{G.~Ditzler}, \bibinfo{author}{M.~Roveri}, \bibinfo{author}{C.~Alippi}, \bibinfo{author}{R.~Polikar},
\newblock \bibinfo{title}{Learning in nonstationary environments: A survey},
\newblock \bibinfo{journal}{IEEE Computational Intelligence Magazine} \bibinfo{volume}{10} (\bibinfo{year}{2015}) \bibinfo{pages}{12--25}.
\bibitem[{Ring(1994)}]{ring_continual_1994}
\bibinfo{author}{M.~Ring}, \bibinfo{title}{Continual learning in reinforcement environments}, \bibinfo{year}{1994}. \URLprefix \url{https://www.proquest.com/openview/2d2f13eb52fc09d3eadfd0c81fe5f181/1?cbl=18750&diss=y&pq-origsite=gscholar}.
\bibitem[{Kumar and III(2012)}]{kumar_2012}
\bibinfo{author}{A.~Kumar}, \bibinfo{author}{H.~D. III}, \bibinfo{title}{Learning {Task} {Grouping} and {Overlap} in {Multi}-task {Learning}}, \bibinfo{year}{2012}. \URLprefix \url{http://arxiv.org/abs/1206.6417}. \DOIprefix\doi{10.48550/arXiv.1206.6417}, \bibinfo{note}{arXiv:1206.6417 [cs]}.
\bibitem[{Giannini et~al.(2024)Giannini, Ziffer, Cossu, and Lomonaco}]{giannini_2024}
\bibinfo{author}{F.~Giannini}, \bibinfo{author}{G.~Ziffer}, \bibinfo{author}{A.~Cossu}, \bibinfo{author}{V.~Lomonaco},
\newblock \bibinfo{title}{Streaming {Continual} {Learning} for {Unified} {Adaptive} {Intelligence} in {Dynamic} {Environments}},
\newblock \bibinfo{journal}{IEEE Intelligent Systems} \bibinfo{volume}{39} (\bibinfo{year}{2024}) \bibinfo{pages}{81--85}. \URLprefix \url{https://ieeexplore.ieee.org/document/10779199/?arnumber=10779199}. \DOIprefix\doi{10.1109/MIS.2024.3479469}, \bibinfo{note}{conference Name: IEEE Intelligent Systems}.
\bibitem[{Li et~al.(2024)Li, Qi, Yan, Zhang, Lei, Tang, and Luo}]{li_2024}
\bibinfo{author}{W.~Li}, \bibinfo{author}{F.~Qi}, \bibinfo{author}{R.~Yan}, \bibinfo{author}{H.~Zhang}, \bibinfo{author}{W.~Lei}, \bibinfo{author}{J.~Tang}, \bibinfo{author}{J.~Luo},
\newblock \bibinfo{title}{Continual {Learning} meets {Multimodal} {Foundation} {Models}: {Fundamentals} and {Advances}},
\newblock in: \bibinfo{booktitle}{Proceedings of the 1st on {Continual} {Learning} meets {Multimodal} {Foundation} {Models}: {Fundamentals} and {Advances}}, {ACMMM} {CL}'24, \bibinfo{publisher}{Association for Computing Machinery}, \bibinfo{address}{New York, NY, USA}, \bibinfo{year}{2024}, pp. \bibinfo{pages}{1--4}. \URLprefix \url{https://dl.acm.org/doi/10.1145/3688859.3690083}. \DOIprefix\doi{10.1145/3688859.3690083}.
\bibitem[{Cossu et~al.(2024)Cossu, Carta, Passaro, Lomonaco, Tuytelaars, and Bacciu}]{cossu_continual_2024}
\bibinfo{author}{A.~Cossu}, \bibinfo{author}{A.~Carta}, \bibinfo{author}{L.~Passaro}, \bibinfo{author}{V.~Lomonaco}, \bibinfo{author}{T.~Tuytelaars}, \bibinfo{author}{D.~Bacciu},
\newblock \bibinfo{title}{Continual pre-training mitigates forgetting in language and vision},
\newblock \bibinfo{journal}{Neural Networks} \bibinfo{volume}{179} (\bibinfo{year}{2024}) \bibinfo{pages}{106492}.
\bibitem[{Zheng et~al.(2025)Zheng, Shi, Cai, Li, Zhang, Li, Yu, and Ma}]{zheng_2025}
\bibinfo{author}{J.~Zheng}, \bibinfo{author}{C.~Shi}, \bibinfo{author}{X.~Cai}, \bibinfo{author}{Q.~Li}, \bibinfo{author}{D.~Zhang}, \bibinfo{author}{C.~Li}, \bibinfo{author}{D.~Yu}, \bibinfo{author}{Q.~Ma}, \bibinfo{title}{Lifelong {Learning} of {Large} {Language} {Model} based {Agents}: {A} {Roadmap}}, \bibinfo{year}{2025}. \URLprefix \url{http://arxiv.org/abs/2501.07278}. \DOIprefix\doi{10.48550/arXiv.2501.07278}, \bibinfo{note}{arXiv:2501.07278 [cs]}.
\bibitem[{Shi et~al.(2024)Shi, Xu, Wang, Qin, Wang, Wang, Wang, Ebrahimi, and Wang}]{shi_continual_2024}
\bibinfo{author}{H.~Shi}, \bibinfo{author}{Z.~Xu}, \bibinfo{author}{H.~Wang}, \bibinfo{author}{W.~Qin}, \bibinfo{author}{W.~Wang}, \bibinfo{author}{Y.~Wang}, \bibinfo{author}{Z.~Wang}, \bibinfo{author}{S.~Ebrahimi}, \bibinfo{author}{H.~Wang},
\newblock \bibinfo{title}{Continual learning of large language models: A comprehensive survey},
\newblock \bibinfo{journal}{arXiv preprint arXiv:2404.16789}  (\bibinfo{year}{2024}).
\bibitem[{Van~de Ven et~al.(2022)Van~de Ven, Tuytelaars, and Tolias}]{van_de_ven_2022}
\bibinfo{author}{G.~M. Van~de Ven}, \bibinfo{author}{T.~Tuytelaars}, \bibinfo{author}{A.~S. Tolias},
\newblock \bibinfo{title}{Three types of incremental learning},
\newblock \bibinfo{journal}{Nature Machine Intelligence} \bibinfo{volume}{4} (\bibinfo{year}{2022}) \bibinfo{pages}{1185--1197}.
\bibitem[{Hensch(2005)}]{hensch_2005}
\bibinfo{author}{T.~K. Hensch},
\newblock \bibinfo{title}{Critical period plasticity in local cortical circuits},
\newblock \bibinfo{journal}{Nature reviews neuroscience} \bibinfo{volume}{6} (\bibinfo{year}{2005}) \bibinfo{pages}{877--888}.
\bibitem[{Chen et~al.(2025)Chen, Wang, Cao, Liu, Gao, Cui, Zhu, Ye, Tian, Liu, Gu, Wang, Li, Ren, Chen, Luo, Wang, Jiang, Wang, He, Shi, Zhang, Lv, Wang, Shao, Chu, Tu, He, Wu, Deng, Ge, Chen, Zhang, Wang, Dou, Lu, Zhu, Lu, Lin, Qiao, Dai, and Wang}]{chen_expanding_2025}
\bibinfo{author}{Z.~Chen}, \bibinfo{author}{W.~Wang}, \bibinfo{author}{Y.~Cao}, \bibinfo{author}{Y.~Liu}, \bibinfo{author}{Z.~Gao}, \bibinfo{author}{E.~Cui}, \bibinfo{author}{J.~Zhu}, \bibinfo{author}{S.~Ye}, \bibinfo{author}{H.~Tian}, \bibinfo{author}{Z.~Liu}, \bibinfo{author}{L.~Gu}, \bibinfo{author}{X.~Wang}, \bibinfo{author}{Q.~Li}, \bibinfo{author}{Y.~Ren}, \bibinfo{author}{Z.~Chen}, \bibinfo{author}{J.~Luo}, \bibinfo{author}{J.~Wang}, \bibinfo{author}{T.~Jiang}, \bibinfo{author}{B.~Wang}, \bibinfo{author}{C.~He}, \bibinfo{author}{B.~Shi}, \bibinfo{author}{X.~Zhang}, \bibinfo{author}{H.~Lv}, \bibinfo{author}{Y.~Wang}, \bibinfo{author}{W.~Shao}, \bibinfo{author}{P.~Chu}, \bibinfo{author}{Z.~Tu}, \bibinfo{author}{T.~He}, \bibinfo{author}{Z.~Wu}, \bibinfo{author}{H.~Deng}, \bibinfo{author}{J.~Ge}, \bibinfo{author}{K.~Chen}, \bibinfo{author}{K.~Zhang}, \bibinfo{author}{L.~Wang}, \bibinfo{author}{M.~Dou}, \bibinfo{author}{L.~Lu}, \bibinfo{author}{X.~Zhu}, \bibinfo{author}{T.~Lu}, \bibinfo{author}{D.~Lin},
  \bibinfo{author}{Y.~Qiao}, \bibinfo{author}{J.~Dai}, \bibinfo{author}{W.~Wang}, \bibinfo{title}{Expanding {Performance} {Boundaries} of {Open}-{Source} {Multimodal} {Models} with {Model}, {Data}, and {Test}-{Time} {Scaling}}, \bibinfo{year}{2025}. \URLprefix \url{http://arxiv.org/abs/2412.05271}. \DOIprefix\doi{10.48550/arXiv.2412.05271}, \bibinfo{note}{arXiv:2412.05271 [cs]}.
\bibitem[{Schmidgall and Moor(2025)}]{schmidgall_agentrxiv_2025}
\bibinfo{author}{S.~Schmidgall}, \bibinfo{author}{M.~Moor}, \bibinfo{title}{{AgentRxiv}: {Towards} {Collaborative} {Autonomous} {Research}}, \bibinfo{year}{2025}. \URLprefix \url{http://arxiv.org/abs/2503.18102}. \DOIprefix\doi{10.48550/arXiv.2503.18102}, \bibinfo{note}{arXiv:2503.18102 [cs]}.
\bibitem[{Cai et~al.(2023)Cai, Chang, and Han}]{cai2023human}
\bibinfo{author}{Z.~Cai}, \bibinfo{author}{B.~Chang}, \bibinfo{author}{W.~Han},
\newblock \bibinfo{title}{Human-in-the-loop through chain-of-thought},
\newblock \bibinfo{journal}{arXiv preprint arXiv:2306.07932}  (\bibinfo{year}{2023}).
\bibitem[{Kumar et~al.(2025)Kumar, Ashraf, Thawakar, Anwer, Cholakkal, Shah, Yang, Torr, Khan, and Khan}]{kumar_2025}
\bibinfo{author}{K.~Kumar}, \bibinfo{author}{T.~Ashraf}, \bibinfo{author}{O.~Thawakar}, \bibinfo{author}{R.~M. Anwer}, \bibinfo{author}{H.~Cholakkal}, \bibinfo{author}{M.~Shah}, \bibinfo{author}{M.-H. Yang}, \bibinfo{author}{P.~H.~S. Torr}, \bibinfo{author}{F.~S. Khan}, \bibinfo{author}{S.~Khan}, \bibinfo{title}{{LLM} {Post}-{Training}: {A} {Deep} {Dive} into {Reasoning} {Large} {Language} {Models}}, \bibinfo{year}{2025}. \URLprefix \url{http://arxiv.org/abs/2502.21321}. \DOIprefix\doi{10.48550/arXiv.2502.21321}, \bibinfo{note}{arXiv:2502.21321 [cs]}.
\bibitem[{Team et~al.(2024)Team, Georgiev, Lei, Burnell, Bai, Gulati, Tanzer, Vincent, Pan, Wang et~al.}]{team2024gemini}
\bibinfo{author}{G.~Team}, \bibinfo{author}{P.~Georgiev}, \bibinfo{author}{V.~I. Lei}, \bibinfo{author}{R.~Burnell}, \bibinfo{author}{L.~Bai}, \bibinfo{author}{A.~Gulati}, \bibinfo{author}{G.~Tanzer}, \bibinfo{author}{D.~Vincent}, \bibinfo{author}{Z.~Pan}, \bibinfo{author}{S.~Wang}, et~al., \bibinfo{title}{Gemini 1.5: Unlocking multimodal understanding across millions of tokens of context}, \bibinfo{year}{2024}.
\bibitem[{Hong and Page(2004)}]{hong2004groups}
\bibinfo{author}{L.~Hong}, \bibinfo{author}{S.~E. Page},
\newblock \bibinfo{title}{Groups of diverse problem solvers can outperform groups of high-ability problem solvers},
\newblock \bibinfo{journal}{Proceedings of the National Academy of Sciences} \bibinfo{volume}{101} (\bibinfo{year}{2004}) \bibinfo{pages}{16385--16389}.
\bibitem[{Lesort et~al.(2019)Lesort, Lomonaco, Stoian, Maltoni, Filliat, and Díaz-Rodríguez}]{lesort_continual_2019}
\bibinfo{author}{T.~Lesort}, \bibinfo{author}{V.~Lomonaco}, \bibinfo{author}{A.~Stoian}, \bibinfo{author}{D.~Maltoni}, \bibinfo{author}{D.~Filliat}, \bibinfo{author}{N.~Díaz-Rodríguez}, \bibinfo{title}{Continual {Learning} for {Robotics}: {Definition}, {Framework}, {Learning} {Strategies}, {Opportunities} and {Challenges}}, \bibinfo{year}{2019}. \URLprefix \url{http://arxiv.org/abs/1907.00182}. \DOIprefix\doi{10.48550/arXiv.1907.00182}, \bibinfo{note}{arXiv:1907.00182 [cs]}.
\bibitem[{McCloskey and Cohen(1989)}]{mccloskey_catastrophic_1989}
\bibinfo{author}{M.~McCloskey}, \bibinfo{author}{N.~J. Cohen},
\newblock \bibinfo{title}{Catastrophic {Interference} in {Connectionist} {Networks}: {The} {Sequential} {Learning} {Problem}},
\newblock in: \bibinfo{editor}{G.~H. Bower} (Ed.), \bibinfo{booktitle}{Psychology of {Learning} and {Motivation}}, volume~\bibinfo{volume}{24}, \bibinfo{publisher}{Academic Press}, \bibinfo{year}{1989}, pp. \bibinfo{pages}{109--165}. \URLprefix \url{https://www.sciencedirect.com/science/article/pii/S0079742108605368}. \DOIprefix\doi{10.1016/S0079-7421(08)60536-8}.
\bibitem[{Ratcliff(1990)}]{ratcliff1990connectionist}
\bibinfo{author}{R.~Ratcliff},
\newblock \bibinfo{title}{Connectionist models of recognition memory: constraints imposed by learning and forgetting functions.},
\newblock \bibinfo{journal}{Psychological review} \bibinfo{volume}{97} (\bibinfo{year}{1990}) \bibinfo{pages}{285}.
\bibitem[{Srivastava et~al.(2013)Srivastava, Masci, Kazerounian, Gomez, and Schmidhuber}]{srivastava_compete_2013}
\bibinfo{author}{R.~K. Srivastava}, \bibinfo{author}{J.~Masci}, \bibinfo{author}{S.~Kazerounian}, \bibinfo{author}{F.~Gomez}, \bibinfo{author}{J.~Schmidhuber},
\newblock \bibinfo{title}{Compete to {Compute}},
\newblock in: \bibinfo{editor}{C.~J. Burges}, \bibinfo{editor}{L.~Bottou}, \bibinfo{editor}{M.~Welling}, \bibinfo{editor}{Z.~Ghahramani}, \bibinfo{editor}{K.~Q. Weinberger} (Eds.), \bibinfo{booktitle}{Advances in {Neural} {Information} {Processing} {Systems}}, volume~\bibinfo{volume}{26}, \bibinfo{publisher}{Curran Associates, Inc.}, \bibinfo{year}{2013}. \URLprefix \url{https://proceedings.neurips.cc/paper_files/paper/2013/file/8f1d43620bc6bb580df6e80b0dc05c48-Paper.pdf}.
\bibitem[{Goodfellow et~al.(2015)Goodfellow, Mirza, Xiao, Courville, and Bengio}]{goodfellow_empirical_2015}
\bibinfo{author}{I.~J. Goodfellow}, \bibinfo{author}{M.~Mirza}, \bibinfo{author}{D.~Xiao}, \bibinfo{author}{A.~Courville}, \bibinfo{author}{Y.~Bengio}, \bibinfo{title}{An {Empirical} {Investigation} of {Catastrophic} {Forgetting} in {Gradient}-{Based} {Neural} {Networks}}, \bibinfo{year}{2015}. \URLprefix \url{http://arxiv.org/abs/1312.6211}. \DOIprefix\doi{10.48550/arXiv.1312.6211}, \bibinfo{note}{arXiv:1312.6211 [stat]}.
\bibitem[{Li and Hoiem(2017)}]{li_learning_2017}
\bibinfo{author}{Z.~Li}, \bibinfo{author}{D.~Hoiem}, \bibinfo{title}{Learning without {Forgetting}}, \bibinfo{year}{2017}. \URLprefix \url{http://arxiv.org/abs/1606.09282}. \DOIprefix\doi{10.48550/arXiv.1606.09282}, \bibinfo{note}{arXiv:1606.09282 [cs]}.
\bibitem[{Kirkpatrick et~al.(2017)Kirkpatrick, Pascanu, Rabinowitz, Veness, Desjardins, Rusu, Milan, Quan, Ramalho, Grabska-Barwinska, Hassabis, Clopath, Kumaran, and Hadsell}]{kirkpatrick_overcoming_2017}
\bibinfo{author}{J.~Kirkpatrick}, \bibinfo{author}{R.~Pascanu}, \bibinfo{author}{N.~Rabinowitz}, \bibinfo{author}{J.~Veness}, \bibinfo{author}{G.~Desjardins}, \bibinfo{author}{A.~A. Rusu}, \bibinfo{author}{K.~Milan}, \bibinfo{author}{J.~Quan}, \bibinfo{author}{T.~Ramalho}, \bibinfo{author}{A.~Grabska-Barwinska}, \bibinfo{author}{D.~Hassabis}, \bibinfo{author}{C.~Clopath}, \bibinfo{author}{D.~Kumaran}, \bibinfo{author}{R.~Hadsell},
\newblock \bibinfo{title}{Overcoming catastrophic forgetting in neural networks},
\newblock \bibinfo{journal}{Proceedings of the National Academy of Sciences} \bibinfo{volume}{114} (\bibinfo{year}{2017}) \bibinfo{pages}{3521--3526}. \URLprefix \url{http://arxiv.org/abs/1612.00796}. \DOIprefix\doi{10.1073/pnas.1611835114}, \bibinfo{note}{arXiv:1612.00796 [cs]}.
\bibitem[{Wang et~al.(2024)Wang, Zhang, Su, and Zhu}]{wang2024comprehensive}
\bibinfo{author}{L.~Wang}, \bibinfo{author}{X.~Zhang}, \bibinfo{author}{H.~Su}, \bibinfo{author}{J.~Zhu},
\newblock \bibinfo{title}{A comprehensive survey of continual learning: Theory, method and application},
\newblock \bibinfo{journal}{IEEE Transactions on Pattern Analysis and Machine Intelligence}  (\bibinfo{year}{2024}).
\bibitem[{Wickramasinghe et~al.(2023)Wickramasinghe, Saha, and Roy}]{wickramasinghe2023continual}
\bibinfo{author}{B.~Wickramasinghe}, \bibinfo{author}{G.~Saha}, \bibinfo{author}{K.~Roy},
\newblock \bibinfo{title}{Continual learning: A review of techniques, challenges, and future directions},
\newblock \bibinfo{journal}{IEEE Transactions on Artificial Intelligence} \bibinfo{volume}{5} (\bibinfo{year}{2023}) \bibinfo{pages}{2526--2546}.
\bibitem[{Zenke et~al.(2017)Zenke, Poole, and Ganguli}]{zenke_continual_2017}
\bibinfo{author}{F.~Zenke}, \bibinfo{author}{B.~Poole}, \bibinfo{author}{S.~Ganguli}, \bibinfo{title}{Continual {Learning} {Through} {Synaptic} {Intelligence}}, \bibinfo{year}{2017}. \URLprefix \url{http://arxiv.org/abs/1703.04200}. \DOIprefix\doi{10.48550/arXiv.1703.04200}, \bibinfo{note}{arXiv:1703.04200 [cs]}.
\bibitem[{Rusu et~al.(2022)Rusu, Rabinowitz, Desjardins, Soyer, Kirkpatrick, Kavukcuoglu, Pascanu, and Hadsell}]{rusu_progressive_2022}
\bibinfo{author}{A.~A. Rusu}, \bibinfo{author}{N.~C. Rabinowitz}, \bibinfo{author}{G.~Desjardins}, \bibinfo{author}{H.~Soyer}, \bibinfo{author}{J.~Kirkpatrick}, \bibinfo{author}{K.~Kavukcuoglu}, \bibinfo{author}{R.~Pascanu}, \bibinfo{author}{R.~Hadsell}, \bibinfo{title}{Progressive {Neural} {Networks}}, \bibinfo{year}{2022}. \URLprefix \url{http://arxiv.org/abs/1606.04671}. \DOIprefix\doi{10.48550/arXiv.1606.04671}, \bibinfo{note}{arXiv:1606.04671 [cs]}.
\bibitem[{Mallya and Lazebnik(2018)}]{mallya_packnet_2018}
\bibinfo{author}{A.~Mallya}, \bibinfo{author}{S.~Lazebnik}, \bibinfo{title}{{PackNet}: {Adding} {Multiple} {Tasks} to a {Single} {Network} by {Iterative} {Pruning}}, \bibinfo{year}{2018}. \URLprefix \url{http://arxiv.org/abs/1711.05769}. \DOIprefix\doi{10.48550/arXiv.1711.05769}, \bibinfo{note}{arXiv:1711.05769 [cs]}.
\bibitem[{Mallya et~al.(2018)Mallya, Davis, and Lazebnik}]{mallya_piggyback_2018}
\bibinfo{author}{A.~Mallya}, \bibinfo{author}{D.~Davis}, \bibinfo{author}{S.~Lazebnik}, \bibinfo{title}{Piggyback: {Adapting} a {Single} {Network} to {Multiple} {Tasks} by {Learning} to {Mask} {Weights}}, \bibinfo{year}{2018}. \URLprefix \url{http://arxiv.org/abs/1801.06519}. \DOIprefix\doi{10.48550/arXiv.1801.06519}, \bibinfo{note}{arXiv:1801.06519 [cs]}.
\bibitem[{Frankle and Carbin(2019)}]{frankle_lottery_2019}
\bibinfo{author}{J.~Frankle}, \bibinfo{author}{M.~Carbin}, \bibinfo{title}{The {Lottery} {Ticket} {Hypothesis}: {Finding} {Sparse}, {Trainable} {Neural} {Networks}}, \bibinfo{year}{2019}. \URLprefix \url{http://arxiv.org/abs/1803.03635}. \DOIprefix\doi{10.48550/arXiv.1803.03635}, \bibinfo{note}{arXiv:1803.03635 [cs]}.
\bibitem[{Serrà et~al.(2018)Serrà, Surís, Miron, and Karatzoglou}]{serra_overcoming_2018}
\bibinfo{author}{J.~Serrà}, \bibinfo{author}{D.~Surís}, \bibinfo{author}{M.~Miron}, \bibinfo{author}{A.~Karatzoglou}, \bibinfo{title}{Overcoming catastrophic forgetting with hard attention to the task}, \bibinfo{year}{2018}. \URLprefix \url{http://arxiv.org/abs/1801.01423}. \DOIprefix\doi{10.48550/arXiv.1801.01423}, \bibinfo{note}{arXiv:1801.01423 [cs]}.
\bibitem[{Wortsman et~al.(2020)Wortsman, Ramanujan, Liu, Kembhavi, Rastegari, Yosinski, and Farhadi}]{wortsman_supermasks_2020}
\bibinfo{author}{M.~Wortsman}, \bibinfo{author}{V.~Ramanujan}, \bibinfo{author}{R.~Liu}, \bibinfo{author}{A.~Kembhavi}, \bibinfo{author}{M.~Rastegari}, \bibinfo{author}{J.~Yosinski}, \bibinfo{author}{A.~Farhadi},
\newblock \bibinfo{title}{Supermasks in {Superposition}},
\newblock in: \bibinfo{booktitle}{Advances in {Neural} {Information} {Processing} {Systems}}, volume~\bibinfo{volume}{33}, \bibinfo{publisher}{Curran Associates, Inc.}, \bibinfo{year}{2020}, pp. \bibinfo{pages}{15173--15184}. \URLprefix \url{https://proceedings.neurips.cc/paper/2020/hash/ad1f8bb9b51f023cdc80cf94bb615aa9-Abstract.html}.
\bibitem[{Prabhu et~al.(2020)Prabhu, Torr, and Dokania}]{vedaldi_gdumb_2020}
\bibinfo{author}{A.~Prabhu}, \bibinfo{author}{P.~H.~S. Torr}, \bibinfo{author}{P.~K. Dokania},
\newblock \bibinfo{title}{{GDumb}: {A} {Simple} {Approach} that {Questions} {Our} {Progress} in {Continual} {Learning}},
\newblock in: \bibinfo{editor}{A.~Vedaldi}, \bibinfo{editor}{H.~Bischof}, \bibinfo{editor}{T.~Brox}, \bibinfo{editor}{J.-M. Frahm} (Eds.), \bibinfo{booktitle}{Computer {Vision} – {ECCV} 2020}, volume \bibinfo{volume}{12347}, \bibinfo{publisher}{Springer International Publishing}, \bibinfo{address}{Cham}, \bibinfo{year}{2020}, pp. \bibinfo{pages}{524--540}. \URLprefix \url{https://link.springer.com/10.1007/978-3-030-58536-5_31}. \DOIprefix\doi{10.1007/978-3-030-58536-5_31}, \bibinfo{note}{series Title: Lecture Notes in Computer Science}.
\bibitem[{Lopez-Paz and Ranzato(2022)}]{lopez-paz_gradient_2022}
\bibinfo{author}{D.~Lopez-Paz}, \bibinfo{author}{M.~Ranzato}, \bibinfo{title}{Gradient {Episodic} {Memory} for {Continual} {Learning}}, \bibinfo{year}{2022}. \URLprefix \url{http://arxiv.org/abs/1706.08840}. \DOIprefix\doi{10.48550/arXiv.1706.08840}, \bibinfo{note}{arXiv:1706.08840 [cs]}.
\bibitem[{Rebuffi et~al.(2017)Rebuffi, Kolesnikov, Sperl, and Lampert}]{rebuffi_icarl_2017}
\bibinfo{author}{S.-A. Rebuffi}, \bibinfo{author}{A.~Kolesnikov}, \bibinfo{author}{G.~Sperl}, \bibinfo{author}{C.~H. Lampert}, \bibinfo{title}{{iCaRL}: {Incremental} {Classifier} and {Representation} {Learning}}, \bibinfo{year}{2017}. \URLprefix \url{http://arxiv.org/abs/1611.07725}. \DOIprefix\doi{10.48550/arXiv.1611.07725}, \bibinfo{note}{arXiv:1611.07725 [cs]}.
\bibitem[{Pellegrini et~al.(2020)Pellegrini, Graffieti, Lomonaco, and Maltoni}]{pellegrini_latent_2020}
\bibinfo{author}{L.~Pellegrini}, \bibinfo{author}{G.~Graffieti}, \bibinfo{author}{V.~Lomonaco}, \bibinfo{author}{D.~Maltoni}, \bibinfo{title}{Latent {Replay} for {Real}-{Time} {Continual} {Learning}}, \bibinfo{year}{2020}. \URLprefix \url{http://arxiv.org/abs/1912.01100}. \DOIprefix\doi{10.48550/arXiv.1912.01100}, \bibinfo{note}{arXiv:1912.01100 [cs]}.
\bibitem[{Graffieti et~al.(2023)Graffieti, Maltoni, Pellegrini, and Lomonaco}]{graffieti2023generative}
\bibinfo{author}{G.~Graffieti}, \bibinfo{author}{D.~Maltoni}, \bibinfo{author}{L.~Pellegrini}, \bibinfo{author}{V.~Lomonaco},
\newblock \bibinfo{title}{Generative negative replay for continual learning},
\newblock \bibinfo{journal}{Neural Networks} \bibinfo{volume}{162} (\bibinfo{year}{2023}) \bibinfo{pages}{369--383}.
\bibitem[{Van~de Ven and Tolias(2019)}]{van2019three}
\bibinfo{author}{G.~M. Van~de Ven}, \bibinfo{author}{A.~S. Tolias},
\newblock \bibinfo{title}{Three scenarios for continual learning},
\newblock \bibinfo{journal}{arXiv preprint arXiv:1904.07734}  (\bibinfo{year}{2019}).
\bibitem[{Devlin et~al.(2019)Devlin, Chang, Lee, and Toutanova}]{devlin_bert_2019}
\bibinfo{author}{J.~Devlin}, \bibinfo{author}{M.-W. Chang}, \bibinfo{author}{K.~Lee}, \bibinfo{author}{K.~Toutanova}, \bibinfo{title}{{BERT}: {Pre}-training of {Deep} {Bidirectional} {Transformers} for {Language} {Understanding}}, \bibinfo{year}{2019}. \URLprefix \url{http://arxiv.org/abs/1810.04805}. \DOIprefix\doi{10.48550/arXiv.1810.04805}, \bibinfo{note}{arXiv:1810.04805 [cs]}.
\bibitem[{Brown et~al.(2020)Brown, Mann, Ryder, Subbiah, Kaplan, Dhariwal, Neelakantan, Shyam, Sastry, Askell, Agarwal, Herbert-Voss, Krueger, Henighan, Child, Ramesh, Ziegler, Wu, Winter, Hesse, Chen, Sigler, Litwin, Gray, Chess, Clark, Berner, McCandlish, Radford, Sutskever, and Amodei}]{brown_language_2020}
\bibinfo{author}{T.~B. Brown}, \bibinfo{author}{B.~Mann}, \bibinfo{author}{N.~Ryder}, \bibinfo{author}{M.~Subbiah}, \bibinfo{author}{J.~Kaplan}, \bibinfo{author}{P.~Dhariwal}, \bibinfo{author}{A.~Neelakantan}, \bibinfo{author}{P.~Shyam}, \bibinfo{author}{G.~Sastry}, \bibinfo{author}{A.~Askell}, \bibinfo{author}{S.~Agarwal}, \bibinfo{author}{A.~Herbert-Voss}, \bibinfo{author}{G.~Krueger}, \bibinfo{author}{T.~Henighan}, \bibinfo{author}{R.~Child}, \bibinfo{author}{A.~Ramesh}, \bibinfo{author}{D.~M. Ziegler}, \bibinfo{author}{J.~Wu}, \bibinfo{author}{C.~Winter}, \bibinfo{author}{C.~Hesse}, \bibinfo{author}{M.~Chen}, \bibinfo{author}{E.~Sigler}, \bibinfo{author}{M.~Litwin}, \bibinfo{author}{S.~Gray}, \bibinfo{author}{B.~Chess}, \bibinfo{author}{J.~Clark}, \bibinfo{author}{C.~Berner}, \bibinfo{author}{S.~McCandlish}, \bibinfo{author}{A.~Radford}, \bibinfo{author}{I.~Sutskever}, \bibinfo{author}{D.~Amodei}, \bibinfo{title}{Language {Models} are {Few}-{Shot} {Learners}}, \bibinfo{year}{2020}. \URLprefix
  \url{http://arxiv.org/abs/2005.14165}. \DOIprefix\doi{10.48550/arXiv.2005.14165}, \bibinfo{note}{arXiv:2005.14165 [cs]}.
\bibitem[{Radford et~al.(2021)Radford, Kim, Hallacy, Ramesh, Goh, Agarwal, Sastry, Askell, Mishkin, Clark, Krueger, and Sutskever}]{radford_learning_2021}
\bibinfo{author}{A.~Radford}, \bibinfo{author}{J.~W. Kim}, \bibinfo{author}{C.~Hallacy}, \bibinfo{author}{A.~Ramesh}, \bibinfo{author}{G.~Goh}, \bibinfo{author}{S.~Agarwal}, \bibinfo{author}{G.~Sastry}, \bibinfo{author}{A.~Askell}, \bibinfo{author}{P.~Mishkin}, \bibinfo{author}{J.~Clark}, \bibinfo{author}{G.~Krueger}, \bibinfo{author}{I.~Sutskever}, \bibinfo{title}{Learning {Transferable} {Visual} {Models} {From} {Natural} {Language} {Supervision}}, \bibinfo{year}{2021}. \URLprefix \url{http://arxiv.org/abs/2103.00020}. \DOIprefix\doi{10.48550/arXiv.2103.00020}, \bibinfo{note}{arXiv:2103.00020 [cs]}.
\bibitem[{Bommasani et~al.(2022)Bommasani, Hudson, Adeli, Altman, Arora, Arx, Bernstein, Bohg, Bosselut, Brunskill, Brynjolfsson, Buch, Card, Castellon, Chatterji, Chen, Creel, Davis, Demszky, Donahue, Doumbouya, Durmus, Ermon, Etchemendy, Ethayarajh, Fei-Fei, Finn, Gale, Gillespie, Goel, Goodman, Grossman, Guha, Hashimoto, Henderson, Hewitt, Ho, Hong, Hsu, Huang, Icard, Jain, Jurafsky, Kalluri, Karamcheti, Keeling, Khani, Khattab, Koh, Krass, Krishna, Kuditipudi, Kumar, Ladhak, Lee, Lee, Leskovec, Levent, Li, Li, Ma, Malik, Manning, Mirchandani, Mitchell, Munyikwa, Nair, Narayan, Narayanan, Newman, Nie, Niebles, Nilforoshan, Nyarko, Ogut, Orr, Papadimitriou, Park, Piech, Portelance, Potts, Raghunathan, Reich, Ren, Rong, Roohani, Ruiz, Ryan, Ré, Sadigh, Sagawa, Santhanam, Shih, Srinivasan, Tamkin, Taori, Thomas, Tramèr, Wang, Wang, Wu, Wu, Wu, Xie, Yasunaga, You, Zaharia, Zhang, Zhang, Zhang, Zhang, Zheng, Zhou, and Liang}]{bommasani_opportunities_2022}
\bibinfo{author}{R.~Bommasani}, \bibinfo{author}{D.~A. Hudson}, \bibinfo{author}{E.~Adeli}, \bibinfo{author}{R.~Altman}, \bibinfo{author}{S.~Arora}, \bibinfo{author}{S.~v. Arx}, \bibinfo{author}{M.~S. Bernstein}, \bibinfo{author}{J.~Bohg}, \bibinfo{author}{A.~Bosselut}, \bibinfo{author}{E.~Brunskill}, \bibinfo{author}{E.~Brynjolfsson}, \bibinfo{author}{S.~Buch}, \bibinfo{author}{D.~Card}, \bibinfo{author}{R.~Castellon}, \bibinfo{author}{N.~Chatterji}, \bibinfo{author}{A.~Chen}, \bibinfo{author}{K.~Creel}, \bibinfo{author}{J.~Q. Davis}, \bibinfo{author}{D.~Demszky}, \bibinfo{author}{C.~Donahue}, \bibinfo{author}{M.~Doumbouya}, \bibinfo{author}{E.~Durmus}, \bibinfo{author}{S.~Ermon}, \bibinfo{author}{J.~Etchemendy}, \bibinfo{author}{K.~Ethayarajh}, \bibinfo{author}{L.~Fei-Fei}, \bibinfo{author}{C.~Finn}, \bibinfo{author}{T.~Gale}, \bibinfo{author}{L.~Gillespie}, \bibinfo{author}{K.~Goel}, \bibinfo{author}{N.~Goodman}, \bibinfo{author}{S.~Grossman}, \bibinfo{author}{N.~Guha}, \bibinfo{author}{T.~Hashimoto},
  \bibinfo{author}{P.~Henderson}, \bibinfo{author}{J.~Hewitt}, \bibinfo{author}{D.~E. Ho}, \bibinfo{author}{J.~Hong}, \bibinfo{author}{K.~Hsu}, \bibinfo{author}{J.~Huang}, \bibinfo{author}{T.~Icard}, \bibinfo{author}{S.~Jain}, \bibinfo{author}{D.~Jurafsky}, \bibinfo{author}{P.~Kalluri}, \bibinfo{author}{S.~Karamcheti}, \bibinfo{author}{G.~Keeling}, \bibinfo{author}{F.~Khani}, \bibinfo{author}{O.~Khattab}, \bibinfo{author}{P.~W. Koh}, \bibinfo{author}{M.~Krass}, \bibinfo{author}{R.~Krishna}, \bibinfo{author}{R.~Kuditipudi}, \bibinfo{author}{A.~Kumar}, \bibinfo{author}{F.~Ladhak}, \bibinfo{author}{M.~Lee}, \bibinfo{author}{T.~Lee}, \bibinfo{author}{J.~Leskovec}, \bibinfo{author}{I.~Levent}, \bibinfo{author}{X.~L. Li}, \bibinfo{author}{X.~Li}, \bibinfo{author}{T.~Ma}, \bibinfo{author}{A.~Malik}, \bibinfo{author}{C.~D. Manning}, \bibinfo{author}{S.~Mirchandani}, \bibinfo{author}{E.~Mitchell}, \bibinfo{author}{Z.~Munyikwa}, \bibinfo{author}{S.~Nair}, \bibinfo{author}{A.~Narayan}, \bibinfo{author}{D.~Narayanan},
  \bibinfo{author}{B.~Newman}, \bibinfo{author}{A.~Nie}, \bibinfo{author}{J.~C. Niebles}, \bibinfo{author}{H.~Nilforoshan}, \bibinfo{author}{J.~Nyarko}, \bibinfo{author}{G.~Ogut}, \bibinfo{author}{L.~Orr}, \bibinfo{author}{I.~Papadimitriou}, \bibinfo{author}{J.~S. Park}, \bibinfo{author}{C.~Piech}, \bibinfo{author}{E.~Portelance}, \bibinfo{author}{C.~Potts}, \bibinfo{author}{A.~Raghunathan}, \bibinfo{author}{R.~Reich}, \bibinfo{author}{H.~Ren}, \bibinfo{author}{F.~Rong}, \bibinfo{author}{Y.~Roohani}, \bibinfo{author}{C.~Ruiz}, \bibinfo{author}{J.~Ryan}, \bibinfo{author}{C.~Ré}, \bibinfo{author}{D.~Sadigh}, \bibinfo{author}{S.~Sagawa}, \bibinfo{author}{K.~Santhanam}, \bibinfo{author}{A.~Shih}, \bibinfo{author}{K.~Srinivasan}, \bibinfo{author}{A.~Tamkin}, \bibinfo{author}{R.~Taori}, \bibinfo{author}{A.~W. Thomas}, \bibinfo{author}{F.~Tramèr}, \bibinfo{author}{R.~E. Wang}, \bibinfo{author}{W.~Wang}, \bibinfo{author}{B.~Wu}, \bibinfo{author}{J.~Wu}, \bibinfo{author}{Y.~Wu}, \bibinfo{author}{S.~M. Xie},
  \bibinfo{author}{M.~Yasunaga}, \bibinfo{author}{J.~You}, \bibinfo{author}{M.~Zaharia}, \bibinfo{author}{M.~Zhang}, \bibinfo{author}{T.~Zhang}, \bibinfo{author}{X.~Zhang}, \bibinfo{author}{Y.~Zhang}, \bibinfo{author}{L.~Zheng}, \bibinfo{author}{K.~Zhou}, \bibinfo{author}{P.~Liang}, \bibinfo{title}{On the {Opportunities} and {Risks} of {Foundation} {Models}}, \bibinfo{year}{2022}. \URLprefix \url{http://arxiv.org/abs/2108.07258}. \DOIprefix\doi{10.48550/arXiv.2108.07258}, \bibinfo{note}{arXiv:2108.07258 [cs]}.
\bibitem[{Zhou et~al.(2024)Zhou, Li, Li, Yu, Liu, Wang, Zhang, Ji, Yan, He et~al.}]{zhou2024comprehensive}
\bibinfo{author}{C.~Zhou}, \bibinfo{author}{Q.~Li}, \bibinfo{author}{C.~Li}, \bibinfo{author}{J.~Yu}, \bibinfo{author}{Y.~Liu}, \bibinfo{author}{G.~Wang}, \bibinfo{author}{K.~Zhang}, \bibinfo{author}{C.~Ji}, \bibinfo{author}{Q.~Yan}, \bibinfo{author}{L.~He}, et~al.,
\newblock \bibinfo{title}{A comprehensive survey on pretrained foundation models: A history from bert to chatgpt},
\newblock \bibinfo{journal}{International Journal of Machine Learning and Cybernetics}  (\bibinfo{year}{2024}) \bibinfo{pages}{1--65}.
\bibitem[{Yang et~al.(2025)Yang, Zhou, Ding, Huai, Liu, Chen, Xie, and He}]{yang2025recent}
\bibinfo{author}{Y.~Yang}, \bibinfo{author}{J.~Zhou}, \bibinfo{author}{X.~Ding}, \bibinfo{author}{T.~Huai}, \bibinfo{author}{S.~Liu}, \bibinfo{author}{Q.~Chen}, \bibinfo{author}{Y.~Xie}, \bibinfo{author}{L.~He},
\newblock \bibinfo{title}{Recent advances of foundation language models-based continual learning: A survey},
\newblock \bibinfo{journal}{ACM Computing Surveys} \bibinfo{volume}{57} (\bibinfo{year}{2025}) \bibinfo{pages}{1--38}.
\bibitem[{Plaat et~al.(2024)Plaat, Wong, Verberne, Broekens, Stein, and Back}]{plaat_reasoning_2024}
\bibinfo{author}{A.~Plaat}, \bibinfo{author}{A.~Wong}, \bibinfo{author}{S.~Verberne}, \bibinfo{author}{J.~Broekens}, \bibinfo{author}{N.~v. Stein}, \bibinfo{author}{T.~Back}, \bibinfo{title}{Reasoning with {Large} {Language} {Models}, a {Survey}}, \bibinfo{year}{2024}. \URLprefix \url{http://arxiv.org/abs/2407.11511}. \DOIprefix\doi{10.48550/arXiv.2407.11511}, \bibinfo{note}{arXiv:2407.11511 [cs] version: 1}.
\bibitem[{Hoffmann et~al.(2022)Hoffmann, Borgeaud, Mensch, Buchatskaya, Cai, Rutherford, Casas, Hendricks, Welbl, Clark, Hennigan, Noland, Millican, Driessche, Damoc, Guy, Osindero, Simonyan, Elsen, Rae, Vinyals, and Sifre}]{hoffmann_2022}
\bibinfo{author}{J.~Hoffmann}, \bibinfo{author}{S.~Borgeaud}, \bibinfo{author}{A.~Mensch}, \bibinfo{author}{E.~Buchatskaya}, \bibinfo{author}{T.~Cai}, \bibinfo{author}{E.~Rutherford}, \bibinfo{author}{D.~d.~L. Casas}, \bibinfo{author}{L.~A. Hendricks}, \bibinfo{author}{J.~Welbl}, \bibinfo{author}{A.~Clark}, \bibinfo{author}{T.~Hennigan}, \bibinfo{author}{E.~Noland}, \bibinfo{author}{K.~Millican}, \bibinfo{author}{G.~v.~d. Driessche}, \bibinfo{author}{B.~Damoc}, \bibinfo{author}{A.~Guy}, \bibinfo{author}{S.~Osindero}, \bibinfo{author}{K.~Simonyan}, \bibinfo{author}{E.~Elsen}, \bibinfo{author}{J.~W. Rae}, \bibinfo{author}{O.~Vinyals}, \bibinfo{author}{L.~Sifre}, \bibinfo{title}{Training {Compute}-{Optimal} {Large} {Language} {Models}}, \bibinfo{year}{2022}. \URLprefix \url{http://arxiv.org/abs/2203.15556}. \DOIprefix\doi{10.48550/arXiv.2203.15556}, \bibinfo{note}{arXiv:2203.15556 [cs]}.
\bibitem[{Schwartz et~al.(2020)Schwartz, Dodge, Smith, and Etzioni}]{schwartz_2020}
\bibinfo{author}{R.~Schwartz}, \bibinfo{author}{J.~Dodge}, \bibinfo{author}{N.~A. Smith}, \bibinfo{author}{O.~Etzioni},
\newblock \bibinfo{title}{Green {AI}},
\newblock \bibinfo{journal}{Commun. ACM} \bibinfo{volume}{63} (\bibinfo{year}{2020}) \bibinfo{pages}{54--63}. \URLprefix \url{https://dl.acm.org/doi/10.1145/3381831}. \DOIprefix\doi{10.1145/3381831}.
\bibitem[{Strubell et~al.(2020)Strubell, Ganesh, and McCallum}]{strubell_2020}
\bibinfo{author}{E.~Strubell}, \bibinfo{author}{A.~Ganesh}, \bibinfo{author}{A.~McCallum},
\newblock \bibinfo{title}{Energy and {Policy} {Considerations} for {Modern} {Deep} {Learning} {Research}},
\newblock \bibinfo{journal}{Proceedings of the AAAI Conference on Artificial Intelligence} \bibinfo{volume}{34} (\bibinfo{year}{2020}) \bibinfo{pages}{13693--13696}. \URLprefix \url{https://ojs.aaai.org/index.php/AAAI/article/view/7123}. \DOIprefix\doi{10.1609/aaai.v34i09.7123}, \bibinfo{note}{number: 09}.
\bibitem[{Kirk et~al.(2023)Kirk, Vidgen, Röttger, and Hale}]{kirk_2023}
\bibinfo{author}{H.~R. Kirk}, \bibinfo{author}{B.~Vidgen}, \bibinfo{author}{P.~Röttger}, \bibinfo{author}{S.~A. Hale}, \bibinfo{title}{Personalisation within bounds: {A} risk taxonomy and policy framework for the alignment of large language models with personalised feedback}, \bibinfo{year}{2023}. \URLprefix \url{http://arxiv.org/abs/2303.05453}. \DOIprefix\doi{10.48550/arXiv.2303.05453}, \bibinfo{note}{arXiv:2303.05453 [cs]}.
\bibitem[{Sun et~al.(2019)Sun, Wang, Liu, Miller, Efros, and Hardt}]{sun_2019}
\bibinfo{author}{Y.~Sun}, \bibinfo{author}{X.~Wang}, \bibinfo{author}{Z.~Liu}, \bibinfo{author}{J.~Miller}, \bibinfo{author}{A.~A. Efros}, \bibinfo{author}{M.~Hardt},
\newblock \bibinfo{title}{Test-{Time} {Training} for {Out}-of-{Distribution} {Generalization}}  (\bibinfo{year}{2019}). \URLprefix \url{https://openreview.net/forum?id=HyezmlBKwr}.
\bibitem[{Bender et~al.(2021)Bender, Gebru, McMillan-Major, and Shmitchell}]{bender_2021}
\bibinfo{author}{E.~M. Bender}, \bibinfo{author}{T.~Gebru}, \bibinfo{author}{A.~McMillan-Major}, \bibinfo{author}{S.~Shmitchell},
\newblock \bibinfo{title}{On the {Dangers} of {Stochastic} {Parrots}: {Can} {Language} {Models} {Be} {Too} {Big}?},
\newblock in: \bibinfo{booktitle}{Proceedings of the 2021 {ACM} {Conference} on {Fairness}, {Accountability}, and {Transparency}}, {FAccT} '21, \bibinfo{publisher}{Association for Computing Machinery}, \bibinfo{address}{New York, NY, USA}, \bibinfo{year}{2021}, pp. \bibinfo{pages}{610--623}. \URLprefix \url{https://dl.acm.org/doi/10.1145/3442188.3445922}. \DOIprefix\doi{10.1145/3442188.3445922}.
\bibitem[{Pedreschi et~al.(2024)Pedreschi, Pappalardo, Ferragina, Baeza-Yates, Barab{\'a}si, Dignum, Dignum, Eliassi-Rad, Giannotti, Kert{\'e}sz et~al.}]{pedreschi2024human}
\bibinfo{author}{D.~Pedreschi}, \bibinfo{author}{L.~Pappalardo}, \bibinfo{author}{E.~Ferragina}, \bibinfo{author}{R.~Baeza-Yates}, \bibinfo{author}{A.-L. Barab{\'a}si}, \bibinfo{author}{F.~Dignum}, \bibinfo{author}{V.~Dignum}, \bibinfo{author}{T.~Eliassi-Rad}, \bibinfo{author}{F.~Giannotti}, \bibinfo{author}{J.~Kert{\'e}sz}, et~al.,
\newblock \bibinfo{title}{Human-ai coevolution},
\newblock \bibinfo{journal}{Artificial Intelligence}  (\bibinfo{year}{2024}) \bibinfo{pages}{104244}.
\bibitem[{Raheja and Pochhi(2024)}]{raheja_foundation_2024}
\bibinfo{author}{T.~Raheja}, \bibinfo{author}{N.~Pochhi},
\newblock \bibinfo{title}{Foundation models meet continual learning: Recent advances, challenges, and future directions},
\newblock in: \bibinfo{booktitle}{NeurIPS 2024 Workshop on Scalable Continual Learning for Lifelong Foundation Models}, \bibinfo{year}{2024}.
\bibitem[{Shen et~al.(2023)Shen, Song, Tan, Li, Lu, and Zhuang}]{shen_hugginggpt_2023}
\bibinfo{author}{Y.~Shen}, \bibinfo{author}{K.~Song}, \bibinfo{author}{X.~Tan}, \bibinfo{author}{D.~Li}, \bibinfo{author}{W.~Lu}, \bibinfo{author}{Y.~Zhuang}, \bibinfo{title}{{HuggingGPT}: {Solving} {AI} {Tasks} with {ChatGPT} and its {Friends} in {Hugging} {Face}}, \bibinfo{year}{2023}. \URLprefix \url{http://arxiv.org/abs/2303.17580}. \DOIprefix\doi{10.48550/arXiv.2303.17580}, \bibinfo{note}{arXiv:2303.17580 [cs]}.
\bibitem[{Roth et~al.(2024)Roth, Udandarao, Dziadzio, Prabhu, Cherti, Vinyals, H{\'e}naff, Albanie, Bethge, and Akata}]{roth_practitioner_2024}
\bibinfo{author}{K.~Roth}, \bibinfo{author}{V.~Udandarao}, \bibinfo{author}{S.~Dziadzio}, \bibinfo{author}{A.~Prabhu}, \bibinfo{author}{M.~Cherti}, \bibinfo{author}{O.~Vinyals}, \bibinfo{author}{O.~H{\'e}naff}, \bibinfo{author}{S.~Albanie}, \bibinfo{author}{M.~Bethge}, \bibinfo{author}{Z.~Akata},
\newblock \bibinfo{title}{A practitioner's guide to continual multimodal pretraining},
\newblock \bibinfo{journal}{arXiv preprint arXiv:2408.14471}  (\bibinfo{year}{2024}).
\bibitem[{K{\"u}chemann et~al.(2025)K{\"u}chemann, Avila, Dinc, Hortmann, Revenga, Ruf, Stausberg, Steinert, Fischer, Fischer et~al.}]{kuchemann_opportunities_2025}
\bibinfo{author}{S.~K{\"u}chemann}, \bibinfo{author}{K.~E. Avila}, \bibinfo{author}{Y.~Dinc}, \bibinfo{author}{C.~Hortmann}, \bibinfo{author}{N.~Revenga}, \bibinfo{author}{V.~Ruf}, \bibinfo{author}{N.~Stausberg}, \bibinfo{author}{S.~Steinert}, \bibinfo{author}{F.~Fischer}, \bibinfo{author}{M.~Fischer}, et~al.,
\newblock \bibinfo{title}{On opportunities and challenges of large multimodal foundation models in education},
\newblock \bibinfo{journal}{npj Science of Learning} \bibinfo{volume}{10} (\bibinfo{year}{2025}) \bibinfo{pages}{11}.
\bibitem[{Parmar et~al.(2024)Parmar, Satheesh, Patwary, Shoeybi, and Catanzaro}]{parmar_reuse_2024}
\bibinfo{author}{J.~Parmar}, \bibinfo{author}{S.~Satheesh}, \bibinfo{author}{M.~Patwary}, \bibinfo{author}{M.~Shoeybi}, \bibinfo{author}{B.~Catanzaro},
\newblock \bibinfo{title}{Reuse, don't retrain: A recipe for continued pretraining of language models},
\newblock \bibinfo{journal}{arXiv preprint arXiv:2407.07263}  (\bibinfo{year}{2024}).
\bibitem[{Lla(2025)}]{Llama4Herd2025}
\bibinfo{title}{The {{Llama}} 4 herd: {{The}} beginning of a new era of natively multimodal {{AI}} innovation}, \bibinfo{howpublished}{https://ai.meta.com/blog/llama-4-multimodal-intelligence/}, \bibinfo{year}{2025}.
\bibitem[{Golatkar et~al.(2020)Golatkar, Achille, and Soatto}]{golatkar_eternal_2020}
\bibinfo{author}{A.~Golatkar}, \bibinfo{author}{A.~Achille}, \bibinfo{author}{S.~Soatto},
\newblock \bibinfo{title}{Eternal sunshine of the spotless net: Selective forgetting in deep networks},
\newblock in: \bibinfo{booktitle}{Proceedings of the IEEE/CVF conference on computer vision and pattern recognition}, \bibinfo{year}{2020}, pp. \bibinfo{pages}{9304--9312}.
\bibitem[{Wang et~al.(2025)Wang, Zeng, Guo, Wong, and Gottlob}]{wang_selective_2025}
\bibinfo{author}{L.~Wang}, \bibinfo{author}{X.~Zeng}, \bibinfo{author}{J.~Guo}, \bibinfo{author}{K.-F. Wong}, \bibinfo{author}{G.~Gottlob},
\newblock \bibinfo{title}{Selective forgetting: Advancing machine unlearning techniques and evaluation in language models},
\newblock in: \bibinfo{booktitle}{Proceedings of the AAAI Conference on Artificial Intelligence}, volume~\bibinfo{volume}{39}, \bibinfo{year}{2025}, pp. \bibinfo{pages}{843--851}.
\bibitem[{Zhao et~al.(2024)Zhao, Ni, Fan, Wang, Chen, Meng, and Zhang}]{zhao_continual_2024}
\bibinfo{author}{H.~Zhao}, \bibinfo{author}{B.~Ni}, \bibinfo{author}{J.~Fan}, \bibinfo{author}{Y.~Wang}, \bibinfo{author}{Y.~Chen}, \bibinfo{author}{G.~Meng}, \bibinfo{author}{Z.~Zhang},
\newblock \bibinfo{title}{Continual forgetting for pre-trained vision models},
\newblock in: \bibinfo{booktitle}{Proceedings of the IEEE/CVF Conference on Computer Vision and Pattern Recognition}, \bibinfo{year}{2024}, pp. \bibinfo{pages}{28631--28642}.
\bibitem[{Zhu et~al.(2023)Zhu, Cui, Chen, Qin, Yuan, Fu, Deng, Liu, Sun, and Gu}]{zhu_removing_2023}
\bibinfo{author}{B.~Zhu}, \bibinfo{author}{G.~Cui}, \bibinfo{author}{Y.~Chen}, \bibinfo{author}{Y.~Qin}, \bibinfo{author}{L.~Yuan}, \bibinfo{author}{C.~Fu}, \bibinfo{author}{Y.~Deng}, \bibinfo{author}{Z.~Liu}, \bibinfo{author}{M.~Sun}, \bibinfo{author}{M.~Gu},
\newblock \bibinfo{title}{Removing backdoors in pre-trained models by regularized continual pre-training},
\newblock \bibinfo{journal}{Transactions of the Association for Computational Linguistics} \bibinfo{volume}{11} (\bibinfo{year}{2023}) \bibinfo{pages}{1608--1623}.
\bibitem[{Brinner et~al.(2025)Brinner, Mustafa, and Zarrie{\ss}}]{brinner_enhancing_2025}
\bibinfo{author}{M.~Brinner}, \bibinfo{author}{T.~A. Mustafa}, \bibinfo{author}{S.~Zarrie{\ss}},
\newblock \bibinfo{title}{Enhancing domain-specific encoder models with llm-generated data: How to leverage ontologies, and how to do without them},
\newblock \bibinfo{journal}{arXiv preprint arXiv:2503.22006}  (\bibinfo{year}{2025}).
\bibitem[{Li and Lee(2024)}]{li_examining_2024}
\bibinfo{author}{C.-A. Li}, \bibinfo{author}{H.-Y. Lee},
\newblock \bibinfo{title}{Examining forgetting in continual pre-training of aligned large language models},
\newblock \bibinfo{journal}{arXiv preprint arXiv:2401.03129}  (\bibinfo{year}{2024}).
\bibitem[{Mehta et~al.(2023)Mehta, Patil, Chandar, and Strubell}]{mehta_empirical_2023}
\bibinfo{author}{S.~V. Mehta}, \bibinfo{author}{D.~Patil}, \bibinfo{author}{S.~Chandar}, \bibinfo{author}{E.~Strubell},
\newblock \bibinfo{title}{An empirical investigation of the role of pre-training in lifelong learning},
\newblock \bibinfo{journal}{Journal of Machine Learning Research} \bibinfo{volume}{24} (\bibinfo{year}{2023}) \bibinfo{pages}{1--50}.
\bibitem[{Wu et~al.(2024)Wu, Luo, Li, Pan, Vu, and Haffari}]{wu_continual_2024}
\bibinfo{author}{T.~Wu}, \bibinfo{author}{L.~Luo}, \bibinfo{author}{Y.-F. Li}, \bibinfo{author}{S.~Pan}, \bibinfo{author}{T.-T. Vu}, \bibinfo{author}{G.~Haffari},
\newblock \bibinfo{title}{Continual learning for large language models: A survey},
\newblock \bibinfo{journal}{arXiv preprint arXiv:2402.01364}  (\bibinfo{year}{2024}).
\bibitem[{Zhu et~al.(2023)Zhu, Wei, Liang, Zhang, and Zhao}]{zhu_ctp_2023}
\bibinfo{author}{H.~Zhu}, \bibinfo{author}{Y.~Wei}, \bibinfo{author}{X.~Liang}, \bibinfo{author}{C.~Zhang}, \bibinfo{author}{Y.~Zhao},
\newblock \bibinfo{title}{Ctp: Towards vision-language continual pretraining via compatible momentum contrast and topology preservation},
\newblock in: \bibinfo{booktitle}{Proceedings of the IEEE/CVF International Conference on Computer Vision}, \bibinfo{year}{2023}, pp. \bibinfo{pages}{22257--22267}.
\bibitem[{Zhang et~al.(2023)Zhang, Wang, Kang, Chen, and Wei}]{zhang_slca_2023}
\bibinfo{author}{G.~Zhang}, \bibinfo{author}{L.~Wang}, \bibinfo{author}{G.~Kang}, \bibinfo{author}{L.~Chen}, \bibinfo{author}{Y.~Wei},
\newblock \bibinfo{title}{Slca: Slow learner with classifier alignment for continual learning on a pre-trained model},
\newblock in: \bibinfo{booktitle}{Proceedings of the IEEE/CVF International Conference on Computer Vision}, \bibinfo{year}{2023}, pp. \bibinfo{pages}{19148--19158}.
\bibitem[{Y{\i}ld{\i}z et~al.(2024)Y{\i}ld{\i}z, Ravichandran, Sharma, Bethge, and Ermis}]{yildiz_investigating_2024}
\bibinfo{author}{{\c{C}}.~Y{\i}ld{\i}z}, \bibinfo{author}{N.~K. Ravichandran}, \bibinfo{author}{N.~Sharma}, \bibinfo{author}{M.~Bethge}, \bibinfo{author}{B.~Ermis},
\newblock \bibinfo{title}{Investigating continual pretraining in large language models: Insights and implications},
\newblock \bibinfo{journal}{arXiv preprint arXiv:2402.17400}  (\bibinfo{year}{2024}).
\bibitem[{Guo et~al.(2024)Guo, Fu, Zhang, Zhao, and Shen}]{guo_efficient_2024}
\bibinfo{author}{Y.~Guo}, \bibinfo{author}{J.~Fu}, \bibinfo{author}{H.~Zhang}, \bibinfo{author}{D.~Zhao}, \bibinfo{author}{Y.~Shen},
\newblock \bibinfo{title}{Efficient continual pre-training by mitigating the stability gap},
\newblock \bibinfo{journal}{arXiv preprint arXiv:2406.14833}  (\bibinfo{year}{2024}).
\bibitem[{Li et~al.(2025)Li, Peng, Wang, and Zhang}]{li_open_2025}
\bibinfo{author}{X.~Li}, \bibinfo{author}{L.~Peng}, \bibinfo{author}{Y.-P. Wang}, \bibinfo{author}{W.~Zhang},
\newblock \bibinfo{title}{Open challenges and opportunities in federated foundation models towards biomedical healthcare},
\newblock \bibinfo{journal}{BioData Mining} \bibinfo{volume}{18} (\bibinfo{year}{2025}) \bibinfo{pages}{2}.
\bibitem[{Mendieta et~al.(2023)Mendieta, Han, Shi, Zhu, and Chen}]{mendieta_towards_2023}
\bibinfo{author}{M.~Mendieta}, \bibinfo{author}{B.~Han}, \bibinfo{author}{X.~Shi}, \bibinfo{author}{Y.~Zhu}, \bibinfo{author}{C.~Chen},
\newblock \bibinfo{title}{Towards geospatial foundation models via continual pretraining},
\newblock in: \bibinfo{booktitle}{Proceedings of the IEEE/CVF International Conference on Computer Vision}, \bibinfo{year}{2023}, pp. \bibinfo{pages}{16806--16816}.
\bibitem[{Noce et~al.(2024)Noce, Resta, and Bacciu}]{dalla_sequential_2024}
\bibinfo{author}{N.~D. Noce}, \bibinfo{author}{M.~Resta}, \bibinfo{author}{D.~Bacciu},
\newblock \bibinfo{title}{Sequential continual pre-training for neural machine translation},
\newblock in: \bibinfo{booktitle}{32nd European Symposium on Artificial Neural Networks, Computational Intelligence and Machine Learning, {ESANN}}, \bibinfo{year}{2024}. \URLprefix \url{https://doi.org/10.14428/esann/2024.ES2024-165}.
\bibitem[{Xie et~al.(2024)Xie, Aggarwal, and Ahmad}]{xie_efficient_2024}
\bibinfo{author}{Y.~Xie}, \bibinfo{author}{K.~Aggarwal}, \bibinfo{author}{A.~Ahmad},
\newblock \bibinfo{title}{Efficient continual pre-training for building domain specific large language models},
\newblock in: \bibinfo{booktitle}{Findings of the Association for Computational Linguistics ACL 2024}, \bibinfo{year}{2024}, pp. \bibinfo{pages}{10184--10201}.
\bibitem[{Ostapenko et~al.(2022)Ostapenko, Lesort, Rodriguez, Arefin, Douillard, Rish, and Charlin}]{ostapenko_continual_2022}
\bibinfo{author}{O.~Ostapenko}, \bibinfo{author}{T.~Lesort}, \bibinfo{author}{P.~Rodriguez}, \bibinfo{author}{M.~R. Arefin}, \bibinfo{author}{A.~Douillard}, \bibinfo{author}{I.~Rish}, \bibinfo{author}{L.~Charlin},
\newblock \bibinfo{title}{Continual learning with foundation models: An empirical study of latent replay},
\newblock in: \bibinfo{booktitle}{Conference on lifelong learning agents}, \bibinfo{organization}{PMLR}, \bibinfo{year}{2022}, pp. \bibinfo{pages}{60--91}.
\bibitem[{Houlsby et~al.(2019)Houlsby, Giurgiu, Jastrzebski, Morrone, de~Laroussilhe, Gesmundo, Attariyan, and Gelly}]{houlsby2019parameterefficienttransferlearningnlp}
\bibinfo{author}{N.~Houlsby}, \bibinfo{author}{A.~Giurgiu}, \bibinfo{author}{S.~Jastrzebski}, \bibinfo{author}{B.~Morrone}, \bibinfo{author}{Q.~de~Laroussilhe}, \bibinfo{author}{A.~Gesmundo}, \bibinfo{author}{M.~Attariyan}, \bibinfo{author}{S.~Gelly}, \bibinfo{title}{Parameter-efficient transfer learning for nlp}, \bibinfo{year}{2019}. \URLprefix \url{https://arxiv.org/abs/1902.00751}. \href{http://arxiv.org/abs/1902.00751}{{\tt arXiv:1902.00751}}.
\bibitem[{Hu et~al.(2021)Hu, Shen, Wallis, Allen-Zhu, Li, Wang, Wang, and Chen}]{hu2021loralowrankadaptationlarge}
\bibinfo{author}{E.~J. Hu}, \bibinfo{author}{Y.~Shen}, \bibinfo{author}{P.~Wallis}, \bibinfo{author}{Z.~Allen-Zhu}, \bibinfo{author}{Y.~Li}, \bibinfo{author}{S.~Wang}, \bibinfo{author}{L.~Wang}, \bibinfo{author}{W.~Chen}, \bibinfo{title}{Lora: Low-rank adaptation of large language models}, \bibinfo{year}{2021}. \URLprefix \url{https://arxiv.org/abs/2106.09685}. \href{http://arxiv.org/abs/2106.09685}{{\tt arXiv:2106.09685}}.
\bibitem[{Aggarwal et~al.(2024)Aggarwal, Damle, Goyal, Lokam, and Sitaram}]{aggarwal2024exploringcontinualfinetuningenhancing}
\bibinfo{author}{D.~Aggarwal}, \bibinfo{author}{S.~Damle}, \bibinfo{author}{N.~Goyal}, \bibinfo{author}{S.~Lokam}, \bibinfo{author}{S.~Sitaram}, \bibinfo{title}{Exploring continual fine-tuning for enhancing language ability in large language model}, \bibinfo{year}{2024}. \URLprefix \url{https://arxiv.org/abs/2410.16006}. \href{http://arxiv.org/abs/2410.16006}{{\tt arXiv:2410.16006}}.
\bibitem[{Beaulieu et~al.(2020)Beaulieu, Frati, Miconi, Lehman, Stanley, Clune, and Cheney}]{beaulieu2020learningcontinuallylearn}
\bibinfo{author}{S.~Beaulieu}, \bibinfo{author}{L.~Frati}, \bibinfo{author}{T.~Miconi}, \bibinfo{author}{J.~Lehman}, \bibinfo{author}{K.~O. Stanley}, \bibinfo{author}{J.~Clune}, \bibinfo{author}{N.~Cheney}, \bibinfo{title}{Learning to continually learn}, \bibinfo{year}{2020}. \URLprefix \url{https://arxiv.org/abs/2002.09571}. \href{http://arxiv.org/abs/2002.09571}{{\tt arXiv:2002.09571}}.
\bibitem[{Zhang et~al.(2025)Zhang, Bai, Yang, and Liang}]{zhang2025cloracontinuallowrankadaptation}
\bibinfo{author}{X.~Zhang}, \bibinfo{author}{L.~Bai}, \bibinfo{author}{X.~Yang}, \bibinfo{author}{J.~Liang}, \bibinfo{title}{C-lora: Continual low-rank adaptation for pre-trained models}, \bibinfo{year}{2025}. \URLprefix \url{https://arxiv.org/abs/2502.17920}. \href{http://arxiv.org/abs/2502.17920}{{\tt arXiv:2502.17920}}.
\bibitem[{Fan et~al.(2023)Fan, Kang, Ma, Chen, Wei, Fan, and Yang}]{fan2023fatellmindustrialgradefederated}
\bibinfo{author}{T.~Fan}, \bibinfo{author}{Y.~Kang}, \bibinfo{author}{G.~Ma}, \bibinfo{author}{W.~Chen}, \bibinfo{author}{W.~Wei}, \bibinfo{author}{L.~Fan}, \bibinfo{author}{Q.~Yang}, \bibinfo{title}{Fate-llm: A industrial grade federated learning framework for large language models}, \bibinfo{year}{2023}. \URLprefix \url{https://arxiv.org/abs/2310.10049}. \href{http://arxiv.org/abs/2310.10049}{{\tt arXiv:2310.10049}}.
\bibitem[{Liu et~al.(2024)Liu, Ren, Jin, Zhang, Zhou, Valduriez, and Dou}]{liu2024fisherinformationbasedefficientcurriculum}
\bibinfo{author}{J.~Liu}, \bibinfo{author}{J.~Ren}, \bibinfo{author}{R.~Jin}, \bibinfo{author}{Z.~Zhang}, \bibinfo{author}{Y.~Zhou}, \bibinfo{author}{P.~Valduriez}, \bibinfo{author}{D.~Dou}, \bibinfo{title}{Fisher information-based efficient curriculum federated learning with large language models}, \bibinfo{year}{2024}. \URLprefix \url{https://arxiv.org/abs/2410.00131}. \href{http://arxiv.org/abs/2410.00131}{{\tt arXiv:2410.00131}}.
\bibitem[{Soutif-Cormerais et~al.(2023)Soutif-Cormerais, Carta, Cossu, Hurtado, Lomonaco, Van~de Weijer, and Hemati}]{soutif2023comprehensive}
\bibinfo{author}{A.~Soutif-Cormerais}, \bibinfo{author}{A.~Carta}, \bibinfo{author}{A.~Cossu}, \bibinfo{author}{J.~Hurtado}, \bibinfo{author}{V.~Lomonaco}, \bibinfo{author}{J.~Van~de Weijer}, \bibinfo{author}{H.~Hemati},
\newblock \bibinfo{title}{A comprehensive empirical evaluation on online continual learning},
\newblock in: \bibinfo{booktitle}{Proceedings of the IEEE/CVF International Conference on Computer Vision}, \bibinfo{year}{2023}, pp. \bibinfo{pages}{3518--3528}.
\bibitem[{Parisi et~al.(2019)Parisi, Kemker, Part, Kanan, and Wermter}]{10.1016/j.neunet.2019.01.012}
\bibinfo{author}{G.~I. Parisi}, \bibinfo{author}{R.~Kemker}, \bibinfo{author}{J.~L. Part}, \bibinfo{author}{C.~Kanan}, \bibinfo{author}{S.~Wermter},
\newblock \bibinfo{title}{Continual lifelong learning with neural networks: A review},
\newblock \bibinfo{journal}{Neural Netw.} \bibinfo{volume}{113} (\bibinfo{year}{2019}) \bibinfo{pages}{54–71}. \URLprefix \url{https://doi.org/10.1016/j.neunet.2019.01.012}. \DOIprefix\doi{10.1016/j.neunet.2019.01.012}.
\bibitem[{Du et~al.(2025)Du, Liu, Zheng, Cao, Nakamura, and Chen}]{du2025privacyfinetuninglargelanguage}
\bibinfo{author}{H.~Du}, \bibinfo{author}{S.~Liu}, \bibinfo{author}{L.~Zheng}, \bibinfo{author}{Y.~Cao}, \bibinfo{author}{A.~Nakamura}, \bibinfo{author}{L.~Chen}, \bibinfo{title}{Privacy in fine-tuning large language models: Attacks, defenses, and future directions}, \bibinfo{year}{2025}. \URLprefix \url{https://arxiv.org/abs/2412.16504}. \href{http://arxiv.org/abs/2412.16504}{{\tt arXiv:2412.16504}}.
\bibitem[{Wang et~al.(2022)Wang, Zhang, Lee, Zhang, Sun, Ren, Su, Perot, Dy, and Pfister}]{wang2022learningpromptcontinuallearning}
\bibinfo{author}{Z.~Wang}, \bibinfo{author}{Z.~Zhang}, \bibinfo{author}{C.-Y. Lee}, \bibinfo{author}{H.~Zhang}, \bibinfo{author}{R.~Sun}, \bibinfo{author}{X.~Ren}, \bibinfo{author}{G.~Su}, \bibinfo{author}{V.~Perot}, \bibinfo{author}{J.~Dy}, \bibinfo{author}{T.~Pfister}, \bibinfo{title}{Learning to prompt for continual learning}, \bibinfo{year}{2022}. \URLprefix \url{https://arxiv.org/abs/2112.08654}. \href{http://arxiv.org/abs/2112.08654}{{\tt arXiv:2112.08654}}.
\bibitem[{Smith et~al.(2023)Smith, Karlinsky, Gutta, Cascante-Bonilla, Kim, Arbelle, Panda, Feris, and Kira}]{smith2023codapromptcontinualdecomposedattentionbased}
\bibinfo{author}{J.~S. Smith}, \bibinfo{author}{L.~Karlinsky}, \bibinfo{author}{V.~Gutta}, \bibinfo{author}{P.~Cascante-Bonilla}, \bibinfo{author}{D.~Kim}, \bibinfo{author}{A.~Arbelle}, \bibinfo{author}{R.~Panda}, \bibinfo{author}{R.~Feris}, \bibinfo{author}{Z.~Kira}, \bibinfo{title}{Coda-prompt: Continual decomposed attention-based prompting for rehearsal-free continual learning}, \bibinfo{year}{2023}. \URLprefix \url{https://arxiv.org/abs/2211.13218}. \href{http://arxiv.org/abs/2211.13218}{{\tt arXiv:2211.13218}}.
\bibitem[{Chen et~al.(2024)Chen, Li, Gazagnadou, Zhuang, Chen, and Lyu}]{chen2024duallowrankadaptationcontinual}
\bibinfo{author}{H.~Chen}, \bibinfo{author}{J.~Li}, \bibinfo{author}{N.~Gazagnadou}, \bibinfo{author}{W.~Zhuang}, \bibinfo{author}{C.~Chen}, \bibinfo{author}{L.~Lyu}, \bibinfo{title}{Dual low-rank adaptation for continual learning with pre-trained models}, \bibinfo{year}{2024}. \URLprefix \url{https://arxiv.org/abs/2411.00623}. \href{http://arxiv.org/abs/2411.00623}{{\tt arXiv:2411.00623}}.
\bibitem[{Zhang et~al.(2023)Zhang, Huang, Zhang, Zou, Zheng, and Wang}]{zhang2023adapterlearningpretrainedfeature}
\bibinfo{author}{W.~Zhang}, \bibinfo{author}{Y.~Huang}, \bibinfo{author}{T.~Zhang}, \bibinfo{author}{Q.~Zou}, \bibinfo{author}{W.-S. Zheng}, \bibinfo{author}{R.~Wang}, \bibinfo{title}{Adapter learning in pretrained feature extractor for continual learning of diseases}, \bibinfo{year}{2023}. \URLprefix \url{https://arxiv.org/abs/2304.09042}. \href{http://arxiv.org/abs/2304.09042}{{\tt arXiv:2304.09042}}.
\bibitem[{Gao et~al.(2024)Gao, Dong, He, Wang, and Gong}]{gao2024promptlearningcontinualadapter}
\bibinfo{author}{X.~Gao}, \bibinfo{author}{S.~Dong}, \bibinfo{author}{Y.~He}, \bibinfo{author}{Q.~Wang}, \bibinfo{author}{Y.~Gong}, \bibinfo{title}{Beyond prompt learning: Continual adapter for efficient rehearsal-free continual learning}, \bibinfo{year}{2024}. \URLprefix \url{https://arxiv.org/abs/2407.10281}. \href{http://arxiv.org/abs/2407.10281}{{\tt arXiv:2407.10281}}.
\bibitem[{Yadav et~al.(2023)Yadav, Tam, Choshen, Raffel, and Bansal}]{yadav2023tiesmergingresolvinginterferencemerging}
\bibinfo{author}{P.~Yadav}, \bibinfo{author}{D.~Tam}, \bibinfo{author}{L.~Choshen}, \bibinfo{author}{C.~Raffel}, \bibinfo{author}{M.~Bansal}, \bibinfo{title}{Ties-merging: Resolving interference when merging models}, \bibinfo{year}{2023}. \URLprefix \url{https://arxiv.org/abs/2306.01708}. \href{http://arxiv.org/abs/2306.01708}{{\tt arXiv:2306.01708}}.
\bibitem[{Yu et~al.(2024)Yu, Yu, Yu, Huang, and Li}]{yu2024languagemodelssupermario}
\bibinfo{author}{L.~Yu}, \bibinfo{author}{B.~Yu}, \bibinfo{author}{H.~Yu}, \bibinfo{author}{F.~Huang}, \bibinfo{author}{Y.~Li}, \bibinfo{title}{Language models are super mario: Absorbing abilities from homologous models as a free lunch}, \bibinfo{year}{2024}. \URLprefix \url{https://arxiv.org/abs/2311.03099}. \href{http://arxiv.org/abs/2311.03099}{{\tt arXiv:2311.03099}}.
\bibitem[{Marczak et~al.(2024)Marczak, Twardowski, Trzciński, and Cygert}]{marczak2024magmaxleveragingmodelmerging}
\bibinfo{author}{D.~Marczak}, \bibinfo{author}{B.~Twardowski}, \bibinfo{author}{T.~Trzciński}, \bibinfo{author}{S.~Cygert}, \bibinfo{title}{Magmax: Leveraging model merging for seamless continual learning}, \bibinfo{year}{2024}. \URLprefix \url{https://arxiv.org/abs/2407.06322}. \href{http://arxiv.org/abs/2407.06322}{{\tt arXiv:2407.06322}}.
\bibitem[{Yang et~al.(2024)Yang, Shen, Wang, Guo, Chen, Wang, and Tao}]{yang2024representationsurgerymultitaskmodel}
\bibinfo{author}{E.~Yang}, \bibinfo{author}{L.~Shen}, \bibinfo{author}{Z.~Wang}, \bibinfo{author}{G.~Guo}, \bibinfo{author}{X.~Chen}, \bibinfo{author}{X.~Wang}, \bibinfo{author}{D.~Tao}, \bibinfo{title}{Representation surgery for multi-task model merging}, \bibinfo{year}{2024}. \URLprefix \url{https://arxiv.org/abs/2402.02705}. \href{http://arxiv.org/abs/2402.02705}{{\tt arXiv:2402.02705}}.
\bibitem[{Coleman et~al.(2024)Coleman, Quarantiello, Hurtado, and Lomonaco}]{coleman2024adaptive}
\bibinfo{author}{E.~N. Coleman}, \bibinfo{author}{L.~Quarantiello}, \bibinfo{author}{J.~Hurtado}, \bibinfo{author}{V.~Lomonaco},
\newblock \bibinfo{title}{Adaptive {L}o{RA} merging for efficient domain incremental learning},
\newblock in: \bibinfo{booktitle}{Adaptive Foundation Models: Evolving AI for Personalized and Efficient Learning}, \bibinfo{year}{2024}. \URLprefix \url{https://openreview.net/forum?id=tlB5eonGEk}.
\bibitem[{Finn et~al.(2017)Finn, Abbeel, and Levine}]{finn2017modelagnosticmetalearningfastadaptation}
\bibinfo{author}{C.~Finn}, \bibinfo{author}{P.~Abbeel}, \bibinfo{author}{S.~Levine}, \bibinfo{title}{Model-agnostic meta-learning for fast adaptation of deep networks}, \bibinfo{year}{2017}. \URLprefix \url{https://arxiv.org/abs/1703.03400}. \href{http://arxiv.org/abs/1703.03400}{{\tt arXiv:1703.03400}}.
\bibitem[{Zhang et~al.(2024)Zhang, Qiang, Somayajula, and Xie}]{zhang2024autoloraautomaticallytuningmatrix}
\bibinfo{author}{R.~Zhang}, \bibinfo{author}{R.~Qiang}, \bibinfo{author}{S.~A. Somayajula}, \bibinfo{author}{P.~Xie}, \bibinfo{title}{Autolora: Automatically tuning matrix ranks in low-rank adaptation based on meta learning}, \bibinfo{year}{2024}. \URLprefix \url{https://arxiv.org/abs/2403.09113}. \href{http://arxiv.org/abs/2403.09113}{{\tt arXiv:2403.09113}}.
\bibitem[{Li et~al.(2025)Li, Zou, Tang, Yu, Li, and Li}]{li-etal-2025-meta}
\bibinfo{author}{W.~Li}, \bibinfo{author}{L.~Zou}, \bibinfo{author}{M.~Tang}, \bibinfo{author}{Q.~Yu}, \bibinfo{author}{W.~Li}, \bibinfo{author}{C.~Li},
\newblock \bibinfo{title}{{META}-{LORA}: Memory-efficient sample reweighting for fine-tuning large language models},
\newblock in: \bibinfo{editor}{O.~Rambow}, \bibinfo{editor}{L.~Wanner}, \bibinfo{editor}{M.~Apidianaki}, \bibinfo{editor}{H.~Al-Khalifa}, \bibinfo{editor}{B.~D. Eugenio}, \bibinfo{editor}{S.~Schockaert} (Eds.), \bibinfo{booktitle}{Proceedings of the 31st International Conference on Computational Linguistics}, \bibinfo{publisher}{Association for Computational Linguistics}, \bibinfo{address}{Abu Dhabi, UAE}, \bibinfo{year}{2025}, pp. \bibinfo{pages}{8504--8517}. \URLprefix \url{https://aclanthology.org/2025.coling-main.568/}.
\bibitem[{Palazzo et~al.(2024)Palazzo, Pennisi, Salanitri, Bellitto, Palazzo, and Spampinato}]{palazzo2024fedrewindrewindingcontinualmodel}
\bibinfo{author}{L.~Palazzo}, \bibinfo{author}{M.~Pennisi}, \bibinfo{author}{F.~P. Salanitri}, \bibinfo{author}{G.~Bellitto}, \bibinfo{author}{S.~Palazzo}, \bibinfo{author}{C.~Spampinato}, \bibinfo{title}{Fedrewind: Rewinding continual model exchange for decentralized federated learning}, \bibinfo{year}{2024}. \URLprefix \url{https://arxiv.org/abs/2411.09842}. \href{http://arxiv.org/abs/2411.09842}{{\tt arXiv:2411.09842}}.
\bibitem[{Chollet(2019)}]{chollet2019github}
\bibinfo{author}{F.~Chollet}, \bibinfo{title}{{Abstraction and Reasoning Corpus for Artificial General Intelligence (ARC-AGI)}}, \bibinfo{howpublished}{\url{https://github.com/fchollet/ARC-AGI}}, \bibinfo{year}{2019}. \URLprefix \url{https://github.com/fchollet/ARC-AGI}.
\bibitem[{Srivastava et~al.(2022)Srivastava, Rastogi, Rao, Shoeb, Abid, Fisch, Brown, Santoro, Gupta, Garriga-Alonso et~al.}]{srivastava2022beyond}
\bibinfo{author}{A.~Srivastava}, \bibinfo{author}{A.~Rastogi}, \bibinfo{author}{A.~Rao}, \bibinfo{author}{A.~A.~M. Shoeb}, \bibinfo{author}{A.~Abid}, \bibinfo{author}{A.~Fisch}, \bibinfo{author}{A.~R. Brown}, \bibinfo{author}{A.~Santoro}, \bibinfo{author}{A.~Gupta}, \bibinfo{author}{A.~Garriga-Alonso}, et~al.,
\newblock \bibinfo{title}{Beyond the imitation game: Quantifying and extrapolating the capabilities of language models},
\newblock \bibinfo{journal}{arXiv preprint arXiv:2206.04615}  (\bibinfo{year}{2022}).
\bibitem[{Luo(2024)}]{luo2024has}
\bibinfo{author}{C.~Luo},
\newblock \bibinfo{title}{Has llm reached the scaling ceiling yet? unified insights into llm regularities and constraints},
\newblock \bibinfo{journal}{arXiv preprint arXiv:2412.16443}  (\bibinfo{year}{2024}).
\bibitem[{Tran et~al.(2025)Tran, Dao, Nguyen, Pham, O'Sullivan, and Nguyen}]{tran2025multi}
\bibinfo{author}{K.-T. Tran}, \bibinfo{author}{D.~Dao}, \bibinfo{author}{M.-D. Nguyen}, \bibinfo{author}{Q.-V. Pham}, \bibinfo{author}{B.~O'Sullivan}, \bibinfo{author}{H.~D. Nguyen},
\newblock \bibinfo{title}{Multi-agent collaboration mechanisms: A survey of llms},
\newblock \bibinfo{journal}{arXiv preprint arXiv:2501.06322}  (\bibinfo{year}{2025}).
\bibitem[{Wang et~al.(2024)Wang, Ma, Feng, Zhang, Yang, Zhang, Chen, Tang, Chen, Lin et~al.}]{wang2024survey}
\bibinfo{author}{L.~Wang}, \bibinfo{author}{C.~Ma}, \bibinfo{author}{X.~Feng}, \bibinfo{author}{Z.~Zhang}, \bibinfo{author}{H.~Yang}, \bibinfo{author}{J.~Zhang}, \bibinfo{author}{Z.~Chen}, \bibinfo{author}{J.~Tang}, \bibinfo{author}{X.~Chen}, \bibinfo{author}{Y.~Lin}, et~al.,
\newblock \bibinfo{title}{A survey on large language model based autonomous agents},
\newblock \bibinfo{journal}{Frontiers of Computer Science} \bibinfo{volume}{18} (\bibinfo{year}{2024}) \bibinfo{pages}{186345}.
\bibitem[{Liu et~al.(2025)Liu, Li, Zhang, Wang, He, Hong, Liu, Zhang, Song, Zhu, Cheng, Wang, Wang, Luo, Jin, Zhang, Liu, Chen, Zhang, Yu, Shi, Li, Wu, Teng, Jia, Xu, Xiang, Lin, Liu, Liu, Su, Sun, Berseth, Nie, Foster, Ward, Wu, Gu, Zhuge, Tang, Wang, You, Wang, Pei, Yang, Qi, and Wu}]{liu2025advanceschallengesfoundationagents}
\bibinfo{author}{B.~Liu}, \bibinfo{author}{X.~Li}, \bibinfo{author}{J.~Zhang}, \bibinfo{author}{J.~Wang}, \bibinfo{author}{T.~He}, \bibinfo{author}{S.~Hong}, \bibinfo{author}{H.~Liu}, \bibinfo{author}{S.~Zhang}, \bibinfo{author}{K.~Song}, \bibinfo{author}{K.~Zhu}, \bibinfo{author}{Y.~Cheng}, \bibinfo{author}{S.~Wang}, \bibinfo{author}{X.~Wang}, \bibinfo{author}{Y.~Luo}, \bibinfo{author}{H.~Jin}, \bibinfo{author}{P.~Zhang}, \bibinfo{author}{O.~Liu}, \bibinfo{author}{J.~Chen}, \bibinfo{author}{H.~Zhang}, \bibinfo{author}{Z.~Yu}, \bibinfo{author}{H.~Shi}, \bibinfo{author}{B.~Li}, \bibinfo{author}{D.~Wu}, \bibinfo{author}{F.~Teng}, \bibinfo{author}{X.~Jia}, \bibinfo{author}{J.~Xu}, \bibinfo{author}{J.~Xiang}, \bibinfo{author}{Y.~Lin}, \bibinfo{author}{T.~Liu}, \bibinfo{author}{T.~Liu}, \bibinfo{author}{Y.~Su}, \bibinfo{author}{H.~Sun}, \bibinfo{author}{G.~Berseth}, \bibinfo{author}{J.~Nie}, \bibinfo{author}{I.~Foster}, \bibinfo{author}{L.~Ward}, \bibinfo{author}{Q.~Wu}, \bibinfo{author}{Y.~Gu},
  \bibinfo{author}{M.~Zhuge}, \bibinfo{author}{X.~Tang}, \bibinfo{author}{H.~Wang}, \bibinfo{author}{J.~You}, \bibinfo{author}{C.~Wang}, \bibinfo{author}{J.~Pei}, \bibinfo{author}{Q.~Yang}, \bibinfo{author}{X.~Qi}, \bibinfo{author}{C.~Wu}, \bibinfo{title}{Advances and challenges in foundation agents: From brain-inspired intelligence to evolutionary, collaborative, and safe systems}, \bibinfo{year}{2025}. \URLprefix \url{https://arxiv.org/abs/2504.01990}. \href{http://arxiv.org/abs/2504.01990}{{\tt arXiv:2504.01990}}.
\bibitem[{Wei et~al.(2022)Wei, Wang, Schuurmans, Bosma, Xia, Chi, Le, Zhou et~al.}]{wei2022chain}
\bibinfo{author}{J.~Wei}, \bibinfo{author}{X.~Wang}, \bibinfo{author}{D.~Schuurmans}, \bibinfo{author}{M.~Bosma}, \bibinfo{author}{F.~Xia}, \bibinfo{author}{E.~Chi}, \bibinfo{author}{Q.~V. Le}, \bibinfo{author}{D.~Zhou}, et~al.,
\newblock \bibinfo{title}{Chain-of-thought prompting elicits reasoning in large language models},
\newblock \bibinfo{journal}{Advances in neural information processing systems} \bibinfo{volume}{35} (\bibinfo{year}{2022}) \bibinfo{pages}{24824--24837}.
\bibitem[{Xi et~al.(2025)Xi, Chen, Guo, He, Ding, Hong, Zhang, Wang, Jin, Zhou et~al.}]{xi2025rise}
\bibinfo{author}{Z.~Xi}, \bibinfo{author}{W.~Chen}, \bibinfo{author}{X.~Guo}, \bibinfo{author}{W.~He}, \bibinfo{author}{Y.~Ding}, \bibinfo{author}{B.~Hong}, \bibinfo{author}{M.~Zhang}, \bibinfo{author}{J.~Wang}, \bibinfo{author}{S.~Jin}, \bibinfo{author}{E.~Zhou}, et~al.,
\newblock \bibinfo{title}{The rise and potential of large language model based agents: A survey},
\newblock \bibinfo{journal}{Science China Information Sciences} \bibinfo{volume}{68} (\bibinfo{year}{2025}) \bibinfo{pages}{121101}.
\bibitem[{ichter et~al.(2023)ichter, Brohan, Chebotar, Finn, Hausman, Herzog, Ho, Ibarz, Irpan, Jang, Julian, Kalashnikov, Levine, Lu, Parada, Rao, Sermanet, Toshev, Vanhoucke, Xia, Xiao, Xu, Yan, Brown, Ahn, Cortes, Sievers, Tan, Xu, Reyes, Rettinghouse, Quiambao, Pastor, Luu, Lee, Kuang, Jesmonth, Joshi, Jeffrey, Ruano, Hsu, Gopalakrishnan, David, Zeng, and Fu}]{ichter2023can}
\bibinfo{author}{b.~ichter}, \bibinfo{author}{A.~Brohan}, \bibinfo{author}{Y.~Chebotar}, \bibinfo{author}{C.~Finn}, \bibinfo{author}{K.~Hausman}, \bibinfo{author}{A.~Herzog}, \bibinfo{author}{D.~Ho}, \bibinfo{author}{J.~Ibarz}, \bibinfo{author}{A.~Irpan}, \bibinfo{author}{E.~Jang}, \bibinfo{author}{R.~Julian}, \bibinfo{author}{D.~Kalashnikov}, \bibinfo{author}{S.~Levine}, \bibinfo{author}{Y.~Lu}, \bibinfo{author}{C.~Parada}, \bibinfo{author}{K.~Rao}, \bibinfo{author}{P.~Sermanet}, \bibinfo{author}{A.~T. Toshev}, \bibinfo{author}{V.~Vanhoucke}, \bibinfo{author}{F.~Xia}, \bibinfo{author}{T.~Xiao}, \bibinfo{author}{P.~Xu}, \bibinfo{author}{M.~Yan}, \bibinfo{author}{N.~Brown}, \bibinfo{author}{M.~Ahn}, \bibinfo{author}{O.~Cortes}, \bibinfo{author}{N.~Sievers}, \bibinfo{author}{C.~Tan}, \bibinfo{author}{S.~Xu}, \bibinfo{author}{D.~Reyes}, \bibinfo{author}{J.~Rettinghouse}, \bibinfo{author}{J.~Quiambao}, \bibinfo{author}{P.~Pastor}, \bibinfo{author}{L.~Luu}, \bibinfo{author}{K.-H. Lee}, \bibinfo{author}{Y.~Kuang},
  \bibinfo{author}{S.~Jesmonth}, \bibinfo{author}{N.~J. Joshi}, \bibinfo{author}{K.~Jeffrey}, \bibinfo{author}{R.~J. Ruano}, \bibinfo{author}{J.~Hsu}, \bibinfo{author}{K.~Gopalakrishnan}, \bibinfo{author}{B.~David}, \bibinfo{author}{A.~Zeng}, \bibinfo{author}{C.~K. Fu},
\newblock \bibinfo{title}{Do as i can, not as i say: Grounding language in robotic affordances},
\newblock in: \bibinfo{editor}{K.~Liu}, \bibinfo{editor}{D.~Kulic}, \bibinfo{editor}{J.~Ichnowski} (Eds.), \bibinfo{booktitle}{Proceedings of The 6th Conference on Robot Learning}, volume \bibinfo{volume}{205} of \textit{\bibinfo{series}{Proceedings of Machine Learning Research}}, \bibinfo{publisher}{PMLR}, \bibinfo{year}{2023}, pp. \bibinfo{pages}{287--318}. \URLprefix \url{https://proceedings.mlr.press/v205/ichter23a.html}.
\bibitem[{Xu et~al.(2023)Xu, Peng, Lei, Mukherjee, Liu, and Xu}]{xu2023rewoo}
\bibinfo{author}{B.~Xu}, \bibinfo{author}{Z.~Peng}, \bibinfo{author}{B.~Lei}, \bibinfo{author}{S.~Mukherjee}, \bibinfo{author}{Y.~Liu}, \bibinfo{author}{D.~Xu},
\newblock \bibinfo{title}{Rewoo: Decoupling reasoning from observations for efficient augmented language models},
\newblock \bibinfo{journal}{CoRR} \bibinfo{volume}{abs/2305.18323} (\bibinfo{year}{2023}). \URLprefix \url{https://doi.org/10.48550/arXiv.2305.18323}.
\bibitem[{Raman et~al.(2022)Raman, Cohen, Rosen, Idrees, Paulius, and Tellex}]{raman2022planning}
\bibinfo{author}{S.~S. Raman}, \bibinfo{author}{V.~Cohen}, \bibinfo{author}{E.~Rosen}, \bibinfo{author}{I.~Idrees}, \bibinfo{author}{D.~Paulius}, \bibinfo{author}{S.~Tellex},
\newblock \bibinfo{title}{Planning with large language models via corrective re-prompting},
\newblock in: \bibinfo{booktitle}{NeurIPS 2022 Foundation Models for Decision Making Workshop}, \bibinfo{year}{2022}. \URLprefix \url{https://openreview.net/forum?id=cMDMRBe1TKs}.
\bibitem[{Yao et~al.(2023)Yao, Yu, Zhao, Shafran, Griffiths, Cao, and Narasimhan}]{yao2023tree}
\bibinfo{author}{S.~Yao}, \bibinfo{author}{D.~Yu}, \bibinfo{author}{J.~Zhao}, \bibinfo{author}{I.~Shafran}, \bibinfo{author}{T.~Griffiths}, \bibinfo{author}{Y.~Cao}, \bibinfo{author}{K.~Narasimhan},
\newblock \bibinfo{title}{Tree of thoughts: Deliberate problem solving with large language models},
\newblock \bibinfo{journal}{Advances in neural information processing systems} \bibinfo{volume}{36} (\bibinfo{year}{2023}) \bibinfo{pages}{11809--11822}.
\bibitem[{Du et~al.(2022)Du, Huang, Dai, Tong, Lepikhin, Xu, Krikun, Zhou, Yu, Firat et~al.}]{du2022glam}
\bibinfo{author}{N.~Du}, \bibinfo{author}{Y.~Huang}, \bibinfo{author}{A.~M. Dai}, \bibinfo{author}{S.~Tong}, \bibinfo{author}{D.~Lepikhin}, \bibinfo{author}{Y.~Xu}, \bibinfo{author}{M.~Krikun}, \bibinfo{author}{Y.~Zhou}, \bibinfo{author}{A.~W. Yu}, \bibinfo{author}{O.~Firat}, et~al.,
\newblock \bibinfo{title}{Glam: Efficient scaling of language models with mixture-of-experts},
\newblock in: \bibinfo{booktitle}{International conference on machine learning}, \bibinfo{organization}{PMLR}, \bibinfo{year}{2022}, pp. \bibinfo{pages}{5547--5569}.
\bibitem[{Lin et~al.(2024)Lin, Tang, Ye, Cui, Zhu, Jin, Huang, Zhang, Pang, Ning et~al.}]{lin2024moe}
\bibinfo{author}{B.~Lin}, \bibinfo{author}{Z.~Tang}, \bibinfo{author}{Y.~Ye}, \bibinfo{author}{J.~Cui}, \bibinfo{author}{B.~Zhu}, \bibinfo{author}{P.~Jin}, \bibinfo{author}{J.~Huang}, \bibinfo{author}{J.~Zhang}, \bibinfo{author}{Y.~Pang}, \bibinfo{author}{M.~Ning}, et~al.,
\newblock \bibinfo{title}{Moe-llava: Mixture of experts for large vision-language models},
\newblock \bibinfo{journal}{arXiv preprint arXiv:2401.15947}  (\bibinfo{year}{2024}).
\bibitem[{Xue et~al.(2024)Xue, Zheng, Fu, Ni, Zheng, Zhou, and You}]{xue2024openmoe}
\bibinfo{author}{F.~Xue}, \bibinfo{author}{Z.~Zheng}, \bibinfo{author}{Y.~Fu}, \bibinfo{author}{J.~Ni}, \bibinfo{author}{Z.~Zheng}, \bibinfo{author}{W.~Zhou}, \bibinfo{author}{Y.~You},
\newblock \bibinfo{title}{Openmoe: An early effort on open mixture-of-experts language models},
\newblock \bibinfo{journal}{arXiv preprint arXiv:2402.01739}  (\bibinfo{year}{2024}).
\bibitem[{Dai et~al.(2024)Dai, Deng, Zhao, Xu, Gao, Chen, Li, Zeng, Yu, Wu et~al.}]{dai2024deepseekmoe}
\bibinfo{author}{D.~Dai}, \bibinfo{author}{C.~Deng}, \bibinfo{author}{C.~Zhao}, \bibinfo{author}{R.~Xu}, \bibinfo{author}{H.~Gao}, \bibinfo{author}{D.~Chen}, \bibinfo{author}{J.~Li}, \bibinfo{author}{W.~Zeng}, \bibinfo{author}{X.~Yu}, \bibinfo{author}{Y.~Wu}, et~al.,
\newblock \bibinfo{title}{Deepseekmoe: Towards ultimate expert specialization in mixture-of-experts language models},
\newblock \bibinfo{journal}{arXiv preprint arXiv:2401.06066}  (\bibinfo{year}{2024}).
\bibitem[{Fischer(2023)}]{fischer2023reflective}
\bibinfo{author}{K.~A. Fischer},
\newblock \bibinfo{title}{Reflective linguistic programming (rlp): A stepping stone in socially-aware agi (socialagi)},
\newblock \bibinfo{journal}{arXiv preprint arXiv:2305.12647}  (\bibinfo{year}{2023}).
\bibitem[{Li et~al.(2023)Li, Wang, Zhu, Zhang, Hou, Lian, and Xie}]{li2023emotionprompt}
\bibinfo{author}{C.~Li}, \bibinfo{author}{J.~Wang}, \bibinfo{author}{K.~Zhu}, \bibinfo{author}{Y.~Zhang}, \bibinfo{author}{W.~Hou}, \bibinfo{author}{J.~Lian}, \bibinfo{author}{X.~Xie},
\newblock \bibinfo{title}{Emotionprompt: Leveraging psychology for large language models enhancement via emotional stimulus},
\newblock \bibinfo{journal}{arXiv preprint arXiv:2307.11760}  (\bibinfo{year}{2023}).
\bibitem[{Wang et~al.(????)Wang, Zhang, Li, Li, Kallidromitis, Kato, Kozuka, and Darrell}]{wangsegllm}
\bibinfo{author}{X.~Wang}, \bibinfo{author}{S.~Zhang}, \bibinfo{author}{S.~Li}, \bibinfo{author}{K.~Li}, \bibinfo{author}{K.~Kallidromitis}, \bibinfo{author}{Y.~Kato}, \bibinfo{author}{K.~Kozuka}, \bibinfo{author}{T.~Darrell},
\newblock \bibinfo{title}{Segllm: Multi-round reasoning segmentation with large language models},
\newblock in: \bibinfo{booktitle}{The Thirteenth International Conference on Learning Representations}, ????
\bibitem[{Zhou et~al.(2025)Zhou, Zhang, Tan, Zhang, and Li}]{zhou2025collaborative}
\bibinfo{author}{Z.~Zhou}, \bibinfo{author}{X.~Zhang}, \bibinfo{author}{S.~Tan}, \bibinfo{author}{L.~Zhang}, \bibinfo{author}{C.~Li},
\newblock \bibinfo{title}{Collaborative evolution: Multi-round learning between large and small language models for emergent fake news detection},
\newblock in: \bibinfo{booktitle}{Proceedings of the AAAI Conference on Artificial Intelligence}, volume~\bibinfo{volume}{39}, \bibinfo{year}{2025}, pp. \bibinfo{pages}{1210--1218}.
\bibitem[{Wu et~al.(2023)Wu, Bansal, Zhang, Wu, Li, Zhu, Jiang, Zhang, Zhang, Liu et~al.}]{wu2023autogen}
\bibinfo{author}{Q.~Wu}, \bibinfo{author}{G.~Bansal}, \bibinfo{author}{J.~Zhang}, \bibinfo{author}{Y.~Wu}, \bibinfo{author}{B.~Li}, \bibinfo{author}{E.~Zhu}, \bibinfo{author}{L.~Jiang}, \bibinfo{author}{X.~Zhang}, \bibinfo{author}{S.~Zhang}, \bibinfo{author}{J.~Liu}, et~al.,
\newblock \bibinfo{title}{Autogen: Enabling next-gen llm applications via multi-agent conversation},
\newblock \bibinfo{journal}{arXiv preprint arXiv:2308.08155}  (\bibinfo{year}{2023}).
\bibitem[{Wang et~al.(2024)Wang, Xie, Jiang, Mandlekar, Xiao, Zhu, Fan, and Anandkumar}]{wang2024voyager}
\bibinfo{author}{G.~Wang}, \bibinfo{author}{Y.~Xie}, \bibinfo{author}{Y.~Jiang}, \bibinfo{author}{A.~Mandlekar}, \bibinfo{author}{C.~Xiao}, \bibinfo{author}{Y.~Zhu}, \bibinfo{author}{L.~Fan}, \bibinfo{author}{A.~Anandkumar},
\newblock \bibinfo{title}{Voyager: An open-ended embodied agent with large language models},
\newblock \bibinfo{journal}{Transactions on Machine Learning Research}  (\bibinfo{year}{2024}). \URLprefix \url{https://openreview.net/forum?id=ehfRiF0R3a}.
\bibitem[{Zhu et~al.(2023)Zhu, Chen, Tian, Tao, Su, Yang, Huang, Li, Lu, Wang, Qiao, Zhang, and Dai}]{zhu2023ghostminecraftgenerallycapable}
\bibinfo{author}{X.~Zhu}, \bibinfo{author}{Y.~Chen}, \bibinfo{author}{H.~Tian}, \bibinfo{author}{C.~Tao}, \bibinfo{author}{W.~Su}, \bibinfo{author}{C.~Yang}, \bibinfo{author}{G.~Huang}, \bibinfo{author}{B.~Li}, \bibinfo{author}{L.~Lu}, \bibinfo{author}{X.~Wang}, \bibinfo{author}{Y.~Qiao}, \bibinfo{author}{Z.~Zhang}, \bibinfo{author}{J.~Dai}, \bibinfo{title}{Ghost in the minecraft: Generally capable agents for open-world environments via large language models with text-based knowledge and memory}, \bibinfo{year}{2023}. \URLprefix \url{https://arxiv.org/abs/2305.17144}. \href{http://arxiv.org/abs/2305.17144}{{\tt arXiv:2305.17144}}.
\bibitem[{Zhang et~al.(2023)Zhang, Yang, Liu, Han, Chen, Huang, Fu, and Yu}]{zhang2023appagent}
\bibinfo{author}{C.~Zhang}, \bibinfo{author}{Z.~Yang}, \bibinfo{author}{J.~Liu}, \bibinfo{author}{Y.~Han}, \bibinfo{author}{X.~Chen}, \bibinfo{author}{Z.~Huang}, \bibinfo{author}{B.~Fu}, \bibinfo{author}{G.~Yu},
\newblock \bibinfo{title}{Appagent: Multimodal agents as smartphone users},
\newblock \bibinfo{journal}{arXiv preprint arXiv:2312.13771}  (\bibinfo{year}{2023}).
\bibitem[{Madaan et~al.(2022)Madaan, Tandon, Clark, and Yang}]{madaan2022memory}
\bibinfo{author}{A.~Madaan}, \bibinfo{author}{N.~Tandon}, \bibinfo{author}{P.~Clark}, \bibinfo{author}{Y.~Yang},
\newblock \bibinfo{title}{Memory-assisted prompt editing to improve gpt-3 after deployment},
\newblock in: \bibinfo{booktitle}{Proceedings of the 2022 Conference on Empirical Methods in Natural Language Processing}, \bibinfo{year}{2022}, pp. \bibinfo{pages}{2833--2861}.
\bibitem[{Sumers et~al.(2024)Sumers, Yao, Narasimhan, and Griffiths}]{sumers2024cognitive}
\bibinfo{author}{T.~Sumers}, \bibinfo{author}{S.~Yao}, \bibinfo{author}{K.~Narasimhan}, \bibinfo{author}{T.~Griffiths},
\newblock \bibinfo{title}{Cognitive architectures for language agents},
\newblock \bibinfo{journal}{Transactions on Machine Learning Research}  (\bibinfo{year}{2024}). \URLprefix \url{https://openreview.net/forum?id=1i6ZCvflQJ}, \bibinfo{note}{survey Certification}.
\bibitem[{Li et~al.(2023)Li, Hammoud, Itani, Khizbullin, and Ghanem}]{li2023camel}
\bibinfo{author}{G.~Li}, \bibinfo{author}{H.~Hammoud}, \bibinfo{author}{H.~Itani}, \bibinfo{author}{D.~Khizbullin}, \bibinfo{author}{B.~Ghanem},
\newblock \bibinfo{title}{Camel: Communicative agents for" mind" exploration of large language model society},
\newblock \bibinfo{journal}{Advances in Neural Information Processing Systems} \bibinfo{volume}{36} (\bibinfo{year}{2023}) \bibinfo{pages}{51991--52008}.
\bibitem[{Liu et~al.(2023)Liu, Zhou, Liu, Zhao, Yao, and Shao}]{liu2023incremental}
\bibinfo{author}{H.~Liu}, \bibinfo{author}{Y.~Zhou}, \bibinfo{author}{B.~Liu}, \bibinfo{author}{J.~Zhao}, \bibinfo{author}{R.~Yao}, \bibinfo{author}{Z.~Shao},
\newblock \bibinfo{title}{Incremental learning with neural networks for computer vision: a survey},
\newblock \bibinfo{journal}{Artificial intelligence review} \bibinfo{volume}{56} (\bibinfo{year}{2023}) \bibinfo{pages}{4557--4589}.
\bibitem[{Khetarpal et~al.(2022)Khetarpal, Riemer, Rish, and Precup}]{khetarpal2022towards}
\bibinfo{author}{K.~Khetarpal}, \bibinfo{author}{M.~Riemer}, \bibinfo{author}{I.~Rish}, \bibinfo{author}{D.~Precup},
\newblock \bibinfo{title}{Towards continual reinforcement learning: A review and perspectives},
\newblock \bibinfo{journal}{Journal of Artificial Intelligence Research} \bibinfo{volume}{75} (\bibinfo{year}{2022}) \bibinfo{pages}{1401--1476}.
\bibitem[{Gou et~al.(2021)Gou, Yu, Maybank, and Tao}]{gou2021knowledge}
\bibinfo{author}{J.~Gou}, \bibinfo{author}{B.~Yu}, \bibinfo{author}{S.~J. Maybank}, \bibinfo{author}{D.~Tao},
\newblock \bibinfo{title}{Knowledge distillation: A survey},
\newblock \bibinfo{journal}{International Journal of Computer Vision} \bibinfo{volume}{129} (\bibinfo{year}{2021}) \bibinfo{pages}{1789--1819}.
\bibitem[{Yang et~al.(2024)Yang, Zhu, Lu, Wang, Chen, Gao, Yan, and Chen}]{yang2024survey}
\bibinfo{author}{C.~Yang}, \bibinfo{author}{Y.~Zhu}, \bibinfo{author}{W.~Lu}, \bibinfo{author}{Y.~Wang}, \bibinfo{author}{Q.~Chen}, \bibinfo{author}{C.~Gao}, \bibinfo{author}{B.~Yan}, \bibinfo{author}{Y.~Chen},
\newblock \bibinfo{title}{Survey on knowledge distillation for large language models: methods, evaluation, and application},
\newblock \bibinfo{journal}{ACM Transactions on Intelligent Systems and Technology}  (\bibinfo{year}{2024}).
\bibitem[{Carta et~al.(2022)Carta, Cossu, Lomonaco, and Bacciu}]{carta2022ex}
\bibinfo{author}{A.~Carta}, \bibinfo{author}{A.~Cossu}, \bibinfo{author}{V.~Lomonaco}, \bibinfo{author}{D.~Bacciu},
\newblock \bibinfo{title}{Ex-model: Continual learning from a stream of trained models},
\newblock in: \bibinfo{booktitle}{Proceedings of the IEEE/CVF conference on computer vision and pattern recognition}, \bibinfo{year}{2022}, pp. \bibinfo{pages}{3790--3799}.
\bibitem[{Kenton et~al.(2021)Kenton, Everitt, Weidinger, Gabriel, Mikulik, and Irving}]{kenton2021alignment}
\bibinfo{author}{Z.~Kenton}, \bibinfo{author}{T.~Everitt}, \bibinfo{author}{L.~Weidinger}, \bibinfo{author}{I.~Gabriel}, \bibinfo{author}{V.~Mikulik}, \bibinfo{author}{G.~Irving},
\newblock \bibinfo{title}{Alignment of language agents},
\newblock \bibinfo{journal}{arXiv preprint arXiv:2103.14659}  (\bibinfo{year}{2021}).
\bibitem[{Du et~al.(2022)Du, Kim, Raheja, Kumar, and Kang}]{du2022read}
\bibinfo{author}{W.~Du}, \bibinfo{author}{Z.~M. Kim}, \bibinfo{author}{V.~Raheja}, \bibinfo{author}{D.~Kumar}, \bibinfo{author}{D.~Kang},
\newblock \bibinfo{title}{Read, revise, repeat: A system demonstration for human-in-the-loop iterative text revision},
\newblock in: \bibinfo{booktitle}{Proceedings of the First Workshop on Intelligent and Interactive Writing Assistants (In2Writing 2022)}, \bibinfo{year}{2022}, pp. \bibinfo{pages}{96--108}.

\end{thebibliography}

\appendix

\end{document}